\documentclass[twocolumn,10pt]{jmlr} % preprint

\usepackage[most]{tcolorbox}
\usepackage{xcolor}

\usepackage{placeins} % for \FloatBarrier


\usepackage{listings}
\definecolor{type1color}{HTML}{A8C0E0}    % Soft Blue - Action/Goal
\definecolor{type2color}{HTML}{A8D8A8}    % Soft Green - Optimization  
\definecolor{type3color}{HTML}{F0DCA0}    % Soft Yellow - Planning

\newtcolorbox{taskbox}[2][]{%
    enhanced,
    breakable,
    colback=#2!5,
    colframe=#2,
    colbacktitle=#2,
    coltitle=black,
    boxrule=0.75pt,
    arc=2pt,
    left=5pt, 
    right=5pt, 
    top=3pt, 
    bottom=5pt,
    toptitle=2pt, 
    bottomtitle=2pt,
    fonttitle=\bfseries\small,
    #1
}

\definecolor{slategray}{HTML}{4A5568}

\newtcolorbox{examplebox}[2][]{%
    enhanced,
    colback=type1color!5,
    colframe=type1color,
    colbacktitle=type1color,
    coltitle=black,
    boxrule=0.75pt,
    arc=2pt,
    left=5pt, 
    right=5pt, 
    top=3pt, 
    bottom=5pt,
    toptitle=2pt, 
    bottomtitle=2pt,
    fonttitle=\bfseries\small,
    #1
}

\usepackage[most]{tcolorbox}
\usepackage{listings}
\usepackage{booktabs}   % for \toprule, \midrule, \bottomrule
\usepackage{graphicx}   % for \resizebox
\usepackage{pifont}     % for \ding{55}
\usepackage{amssymb}    % for \checkmark
\usepackage{multirow}

\newtcolorbox{promptbox}[2][]{%
  enhanced,
  colback=white,
  colframe=black,
  boxrule=0.4pt,
  arc=2pt,
  left=6pt,right=6pt,top=6pt,bottom=6pt,
  title=\textbf{#2},
  fonttitle=\small,
  #1
}

\usepackage{booktabs}
\usepackage{siunitx}
\usepackage{color}
\usepackage{xcolor}
\usepackage{tikz}
\usetikzlibrary{matrix,positioning}
\usepackage[switch]{lineno}

\theorembodyfont{\upshape}
\theoremheaderfont{\scshape}
\theorempostheader{:}
\theoremsep{\newline}

\jmlrvolume{}
\jmlryear{}
\jmlrsubmitted{}
\jmlrpublished{}
\title[ReXSonoVQA]{ReXSonoVQA: A Video QA Benchmark for \linebreak Procedure-Centric Ultrasound Understanding}

\author{%
 \Name{Xucheng Wang, BS}\nametag{\textsuperscript{1}} \Email{davidx\_wang@hms.harvard.edu}\\ [2pt]
 \Name{Xiaoman Zhang, PhD}\nametag{\textsuperscript{1}} \Email{xiaoman\_zhang@hms.harvard.edu}\\ [2pt]
 \Name{Sung Eun Kim, MD}\nametag{\textsuperscript{1}} \Email{sungeun\_kim2@hms.harvard.edu}\\ [2pt]
 \Name{Ankit Pal, BS}\nametag{\textsuperscript{1}} \Email{ankit\_pal@fas.harvard.edu}\\ [2pt]
 \Name{Pranav Rajpurkar, PhD}\nametag{\textsuperscript{1}}\\ [2pt]
 \addr \textsuperscript{1}Department of Biomedical Informatics, Harvard Medical School, Boston, MA
}

\begin{document}
\pagestyle{plain}

\maketitle
\thispagestyle{plain}

\begin{abstract}
Ultrasound acquisition requires skilled probe manipulation and real-time adjustments. Vision-language models (VLMs) could enable autonomous ultrasound systems, but existing benchmarks evaluate only static images, not dynamic procedural understanding. We introduce \textbf{ReXSonoVQA}, a video QA benchmark with \textbf{514} video clips and \textbf{514} questions (249 MCQ, 265 free-response) targeting three competencies: Action-Goal Reasoning, Artifact Resolution \& Optimization, and Procedure Context \& Planning. Zero-shot evaluation of \textcolor{black}{Gemini 3 Pro, Qwen3.5-397B, LLaVA-Video-72B, and Seed 2.0 Pro} shows VLMs can extract some procedural information, but troubleshooting questions remain challenging with minimal gains over text-only baselines, exposing limitations in causal reasoning. ReXSonoVQA enables developing perception systems for ultrasound training, guidance, and robotic automation.
\end{abstract}

\paragraph*{Data and Code Availability}
The dataset comprises publicly available ultrasound demonstration videos sourced from YouTube. \textcolor{black}{We provide YouTube URLs with timestamps and regions of interest, following established practices in video benchmark curation (e.g., HowTo100M \citep{miech2019howto100mlearningtextvideoembedding}, AVOS \citep{10.1001/jamasurg.2023.6262}).} All video frames shown in figures throughout this paper correspond to entries in the dataset; source URLs are provided in the repository. The ReXSonoVQA's dataset and code are publicly available at \href{https://github.com/rajpurkarlab/RexSonoVQA}{https://github.com/rajpurkarlab/RexSonoVQA}.

\paragraph*{Institutional Review Board (IRB)}
 IRB was not required for this study as the data consist exclusively of publicly available demonstration videos sourced from YouTube.

\section{Introduction}
\label{sec:intro}

Ultrasound is widely used because it is portable, safe, and can provide real time imaging, but its acquisition is operator-dependent: obtaining clinically usable views requires skilled, adaptive control of probe position, orientation, pressure, and device settings \citep{10Sharma2021}. This dependence limits both the scalability of ultrasound services and the consistency of scan quality, motivating growing interest in intelligent ultrasound assistance: from real-time guidance systems that support novice operators to fully autonomous robotic scanning \citep{2JIANG2025,9Chen2025, 1JIANG2023, 11Guo2025}. For such systems to succeed, their perception modules must understand not just \textit{what} is visible in static frames, but \textit{how} to obtain and refine it: how probe maneuvers map to imaging goals, how operators troubleshoot artifacts, and how protocols progress.

Vision Language Models (VLMs) have emerged as a promising foundation for such perception systems \citep{12Fung2025,13Salimpour2025} and are increasingly evaluated on medical imaging tasks \citep{4Hu2024,5Chen2024,8Maani2025}. However, existing evaluation has focused almost exclusively on static interpretation spanning from identifying anatomy, recognizing views to detecting pathology from individual images \citep{6Le2025}. This leaves a critical gap: automation systems require temporal and causal reasoning about actions, troubleshooting, and protocol execution that static benchmarks cannot assess \citep{3Munir2025}.

\begin{figure*}[t]
  \centering
  \includegraphics[width=\textwidth]{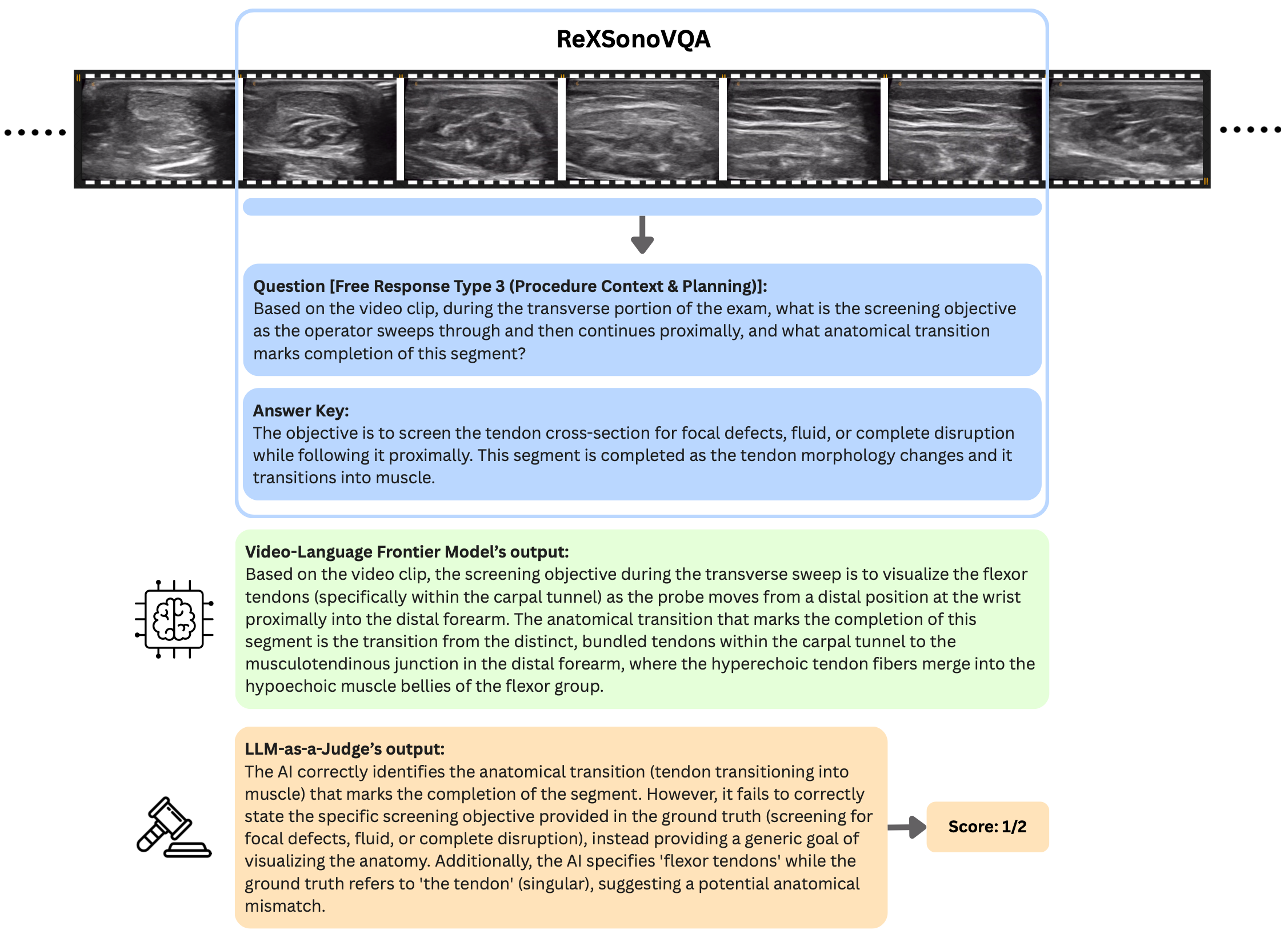}
  \caption{A ReXSonoVQA example: Type 3 (Procedure Context \& Planning, Free-Response) question requiring identification of the screening objective and anatomical transition during a transverse sweep. Gemini 3 Pro correctly identifies the anatomical transition (tendon to muscle) but fails to specify the correct screening objective, receiving a partial score (1/2). More MCQs and free-responses examples see Appendix~\ref{apd:casestudies}.}
  \label{fig:RexSonoExm}
\end{figure*}

\begin{table*}[t]
\centering
\caption{Comparison of ReXSonoVQA with existing medical vision-language benchmarks. \emph{Dynamic Reasoning}: Requires understanding of temporal processes, actions, and causal relationships. \emph{Diagnosis}: Evaluates pathology identification or diagnostic interpretation. \emph{Blind Baseline}: Includes text-only evaluation to verify questions require visual input. \checkmark\ indicates present; \ding{55}\ indicates absent.}
\label{tab:benchmark-comparison}
\footnotesize
\setlength{\tabcolsep}{3pt}
\begin{tabular}{lccccccccc}
\toprule
\textbf{Benchmark} & \textbf{Modality} & \textbf{\# Items} & \textbf{Diagnosis} & \textbf{\shortstack{US-\\Specific}} & \textbf{\shortstack{Multi-\\Organ}} & \textbf{\shortstack{Dynamic\\Reasoning}} & \textbf{MCQ} & \textbf{\shortstack{Free-\\Response}} & \textbf{\shortstack{Blind\\Baseline}} \\
\midrule
MedMCQA         & Text  & 193,155  & \checkmark & \ding{55} & \checkmark & \ding{55} & \checkmark & \ding{55} & \ding{55} \\
GMAI-MMBench    & Image & 26,000   & \checkmark & \ding{55} & \checkmark & \ding{55} & \checkmark & \ding{55} & \ding{55} \\
OmniMedVQA      & Image & 127,995  & \checkmark & \ding{55} & \checkmark & \ding{55} & \checkmark & \ding{55} & \ding{55} \\
VQA-RAD         & Image & 3,515    & \checkmark & \ding{55} & \checkmark & \ding{55} & \checkmark & \checkmark & \ding{55} \\
ReXVQA          & Image & 696,000  & \checkmark & \ding{55} & \ding{55} & \ding{55} & \checkmark & \ding{55} & \ding{55} \\
ReXrank         & Image & 600,000+ & \checkmark & \ding{55} & \ding{55} & \ding{55} & \ding{55} & \checkmark & \ding{55} \\
U2-Bench        & Image & 7,241    & \checkmark & \checkmark & \checkmark & \ding{55} & \checkmark & \checkmark & \ding{55} \\
\midrule
\textbf{ReXSonoVQA} & \textbf{Video} & \textbf{514} & \ding{55} & \checkmark & \checkmark & \checkmark & \checkmark & \checkmark & \checkmark \\
\bottomrule
\end{tabular}
\end{table*}

\begin{figure*}[t]
  \centering
  \includegraphics[width=0.95\textwidth]{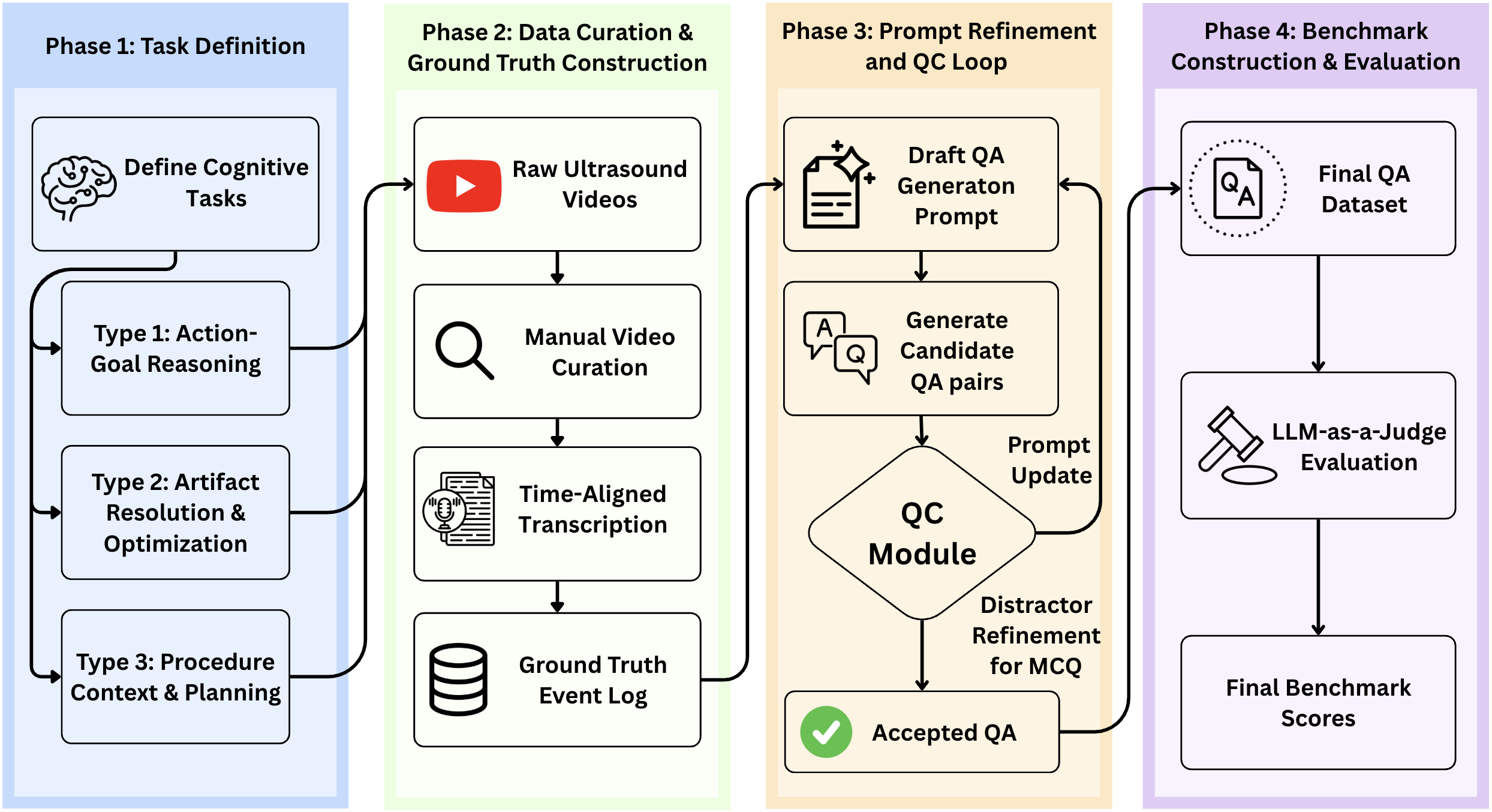}
  \caption{End-to-end pipeline for constructing ReXSonoVQA: (1) Task Definition, (2) Data Curation \& Ground Truth Construction, (3) Prompt Refinement and Quality Control Loop, and (4) Benchmark Construction \& Evaluation.}
  \label{fig:pipeline}
\end{figure*}

We introduce \textbf{ReXSonoVQA}, the first video-based benchmark for procedure-centric ultrasound understanding, comprising 514 video clips paired with \textbf{249} multiple choice (MCQ) and \textbf{265} free-response questions spanning six clinical categories. ReXSonoVQA evaluates dynamic procedural reasoning via three cognitive tasks: (1) \textit{Action--Goal Reasoning}, (2) \textit{Artifact Resolution $\&$ Optimization}, and (3) \textit{Procedure Context $\&$ Planning}. (Fig.~\ref{fig:RexSonoExm}) Our benchmark's construction pipeline leverages instructional videos with timestamped transcripts, applying quality control of QA including blind(text-only) solvability screening and distractor refinement to ensure video is required for answering the question. (Fig.~\ref{fig:pipeline})

While existing VLMs excel at static frame interpretation, dynamic procedural understanding requires frontier video-language models. \textcolor{black}{We evaluate Gemini 3 Pro, Qwen3.5-397B, LLaVA-Video-72B, and Seed 2.0 Pro, all supporting native video input.} Other leading commercial VLMs (GPT-5.2 \citep{openai}, Claude Opus 4.5 \citep{anthropic}) lack native video support and showed poor temporal reasoning with image sequences in exploratory tests. Evaluation under paired conditions (with video vs. text-only) reveals that even with frontier video-language models, substantial gaps remain: Type 2 questions prove most challenging, exposing critical limitations in causal troubleshooting reasoning essential for real-world ultrasound automation.

\section{Related Work}

\paragraph{Vision-Language Models in Medical Imaging.} 
Medical VLMs have advanced medical image interpretation by integrating visual and textual understanding into unified frameworks \citep{li2023llavamed, pmlr-v225-moor23a, Nath_2025_CVPR, Zhang_2024, wu2023generalistfoundationmodelradiology}. Recent ultrasound-specific models, including FetalCLIP \citep{8Maani2025} and LLAUS \citep{11Guo2025}, have demonstrated strong performance in aligning ultrasound features with clinical text 
and performing static anatomy identification. However, existing medical VLMs focus primarily on static, single-frame interpretation, such as identifying anatomy, recognizing views, or detecting pathology from individual images, rather than reasoning about dynamic procedural content. Moreover, most leading commercial VLMs including GPT-5.2 \citep{openai} and Claude Opus 4.5 \citep{anthropic} do not support native video input. Exploratory evaluation using sequential image inputs showed these models failed to perform temporal reasoning effectively (Fig.~\ref{fig:explore}, Appendix~\ref{apd:casestudies}), motivating our \textcolor{black}{focus on models with native video understanding (Gemini 3 Pro, Qwen3.5-397B, LLaVA-Video-72B, and Seed 2.0 Pro)} and extended context windows necessary for analyzing procedural ultrasound sequences.

% VLMs have transformed medical image analysis by automating the translation of visual features to clinical text. 
% Early medical VLMs adapted general-purpose models via domain-specific pre-training, including LLaVA-Med \citep{li2023llavamed} and Med-Flamingo \citep{pmlr-v225-moor23a} for biomedical vocabulary alignment, and VILA-M3 \citep{Nath_2025_CVPR} for reducing hallucinations through expert knowledge incorporation. Within ultrasound, models like FetalCLIP \citep{8Maani2025} and LLAUS \citep{11Guo2025} have demonstrated capabilities in aligning ultrasound features with clinical text and performing static anatomy identification.

% Recent work increasingly evaluates commercial models (GPT, Claude, Gemini) across medical specialties, finding that while they possess broad medical knowledge, they struggle with the precision required for specialized diagnostics \citep{Jiang2024.04.12.24305744}. 
% We select Gemini 3 Pro as our only baseline for its state-of-the-art native video understanding capability and extended context window necessary for analyzing procedural ultrasound sequences. Other leading commercial VLMs, including GPT-5.2 from OpenAI \citep{openai} and Claude Opus 4.5 from Anthropic \citep{anthropic}, do not currently support native video input through their API endpoints.

\paragraph{Ultrasound Automation and Robotics.}
Ultrasound's operator dependence has driven research toward robotic standardization \citep{3Munir2025}. Recent work has shifted from rule-based visual servoing toward learning-based autonomy: UltraBot employs imitation learning for autonomous carotid scanning \citep{2JIANG2025}, USPilot uses LLM-enhanced graph planning to translate natural language into robotic actions \citep{9Chen2025}, and Sonomate provides real-time scanning guidance through visually grounded language models \citep{7Guo2026}. 
Despite this progress, a fundamental challenge persists: current perception systems lack the deep procedural understanding required for adaptive, real-world deployment. Successful automation demands more than anatomy recognition but requires models reason about probe manipulation strategies, diagnose and resolve image quality issues in real-time, adapt protocols to patient-specific constraints, and make decisions about scan progression. These capabilities remain largely unevaluated in existing benchmarks.

\paragraph{Medical AI Benchmarks.} 
Current medical AI benchmarks fall into two main categories: identification/classification and reasoning/generation tasks. Text-centric benchmarks like MedMCQA evaluate professional text-based medical knowledge \citep{pmlr-v174-pal22a}, while visual benchmarks such as OmniMedVQA \citep{4Hu2024}, ReXVQA \citep{pal2025rexvqalargescalevisualquestion}, GMAI-MMBench \citep{5Chen2024}, and VQA-RAD \citep{lau2018dataset} test fine-grained recognition across imaging modalities and anatomical regions. For deeper clinical understanding, benchmarks like ReXrank assess radiology report generation \citep{pmlr-v281-zhang25b}, and U2-Bench evaluates ultrasound image comprehension \citep{6Le2025}.
However, as shown in Table~\ref{tab:benchmark-comparison}, almost all existing benchmarks focus on static image understanding, identifying what is visible in single frames, rather than dynamic reasoning about the acquisition process itself. 
This limitation is particularly problematic for ultrasound automation, where perception systems must understand not just anatomical content but the real-time procedural logic of obtaining and optimizing views. ReXSonoVQA addresses this gap by evaluating procedure-centric reasoning of video sequences. We assess how models understand scanning maneuvers, troubleshooting decisions, protocol progression, and the causal relationships between actions and image outcomes.

\section{Method}
\label{sec:method}

We present a pipeline for constructing a procedure-centric ultrasound video question-answering benchmark that evaluates vision-language models' ability to reason about scan execution: probe maneuvers, acquisition goals, troubleshooting adjustments, and protocol progression. As summarized in Fig.~\ref{fig:pipeline}, the pipeline comprises four stages: (i) task definition, (ii) data curation and ground truth construction, (iii) prompt refinement and quality control loop, and (iv) benchmark construction.

\subsection{Task Definition}
\label{sec:tasks}

Ultrasound acquisition follows a recurring \textit{scan--adjust--progress} loop: operators execute maneuvers toward imaging goals, troubleshoot when quality degrades, and advance through protocol steps. We define three task types that mirror this loop and capture the core competencies required for autonomous ultrasound systems.

% \begin{itemize}
%   \item \textbf{Type 1: Action--Goal Reasoning}\\
%   \emph{Definition:} Identify the probe maneuver and the immediate acquisition goal (target view/region and anatomical location).\\
%   \emph{Example prompt:} ``Based on the clip, what maneuver is performed, and what view/region is the operator trying to acquire?''\\
%   \emph{Targets:} Action recognition and intent grounding (maneuver $\rightarrow$ acquisition target).

%   \item \textbf{Type 2: Artifact Resolution \& Optimization}\\
%   \emph{Definition:} Identify what limits interpretation (poor clarity, target loss, artifact, or structural ambiguity) and what corrective adjustment is made (probe manipulation, patient repositioning, or device settings).\\
%   \emph{Example prompt:} ``The clip shows reduced clarity or ambiguity. What adjustment is made, and what issue does it resolve?''\\
%   \emph{Targets:} Causal reasoning: limitation $\rightarrow$ corrective action $\rightarrow$ expected improvement.

%   \item \textbf{Type 3: Procedure Context \& Planning}\\
%   \emph{Definition:} Infer the operator's position within the protocol (which window/view was just completed) and the next planned step.\\
%   \emph{Example prompt:} ``Based on this segment, what protocol window/view has just been completed, and what should the operator move to next?''\\
%   \emph{Targets:} Workflow understanding: situating local actions within global protocol progression.
% \end{itemize}

\begin{taskbox}[title=Type 1: Action--Goal Reasoning]{type1color}
\small
\textbf{Task:} Identify the probe maneuver and its immediate acquisition goal (target view/region).

\textbf{Motivation:} Autonomous systems must map visual observations to operator intent, the foundation for imitation learning and real-time guidance.

\textbf{Example:} \textit{``Based on the clip, what maneuver is performed, and what view/region is the operator trying to acquire?''}
\end{taskbox}

\vspace{0.5em}

\begin{taskbox}[title=Type 2: Artifact Resolution \& Optimization]{type2color}
\small
\textbf{Task:} Identify what limits image quality (artifact, occlusion, or ambiguity) and what corrective adjustment resolves it.

\textbf{Motivation:} Real-world scanning requires continuous troubleshooting; systems must recognize degradation and select appropriate corrections.

\textbf{Example:} \textit{``The clip shows reduced clarity or ambiguity. What adjustment is made, and what issue does it resolve?''}
\end{taskbox}

\vspace{0.5em}

\begin{taskbox}[title=Type 3: Procedure Context \& Planning]{type3color}
\small
\textbf{Task:} Infer the current position within the protocol and predict the next step.

\textbf{Motivation:} Complete autonomy requires situating local actions within global workflow—knowing what was done and what comes next.

\textbf{Example:} \textit{``Based on this segment, what protocol window/view has just been completed, and what should the operator move to next?''}
\end{taskbox}

\subsection{Data Curation and Ground Truth Construction}
\label{sec:data}

\begin{figure}[t]
  \centering
  \includegraphics[width=\columnwidth]{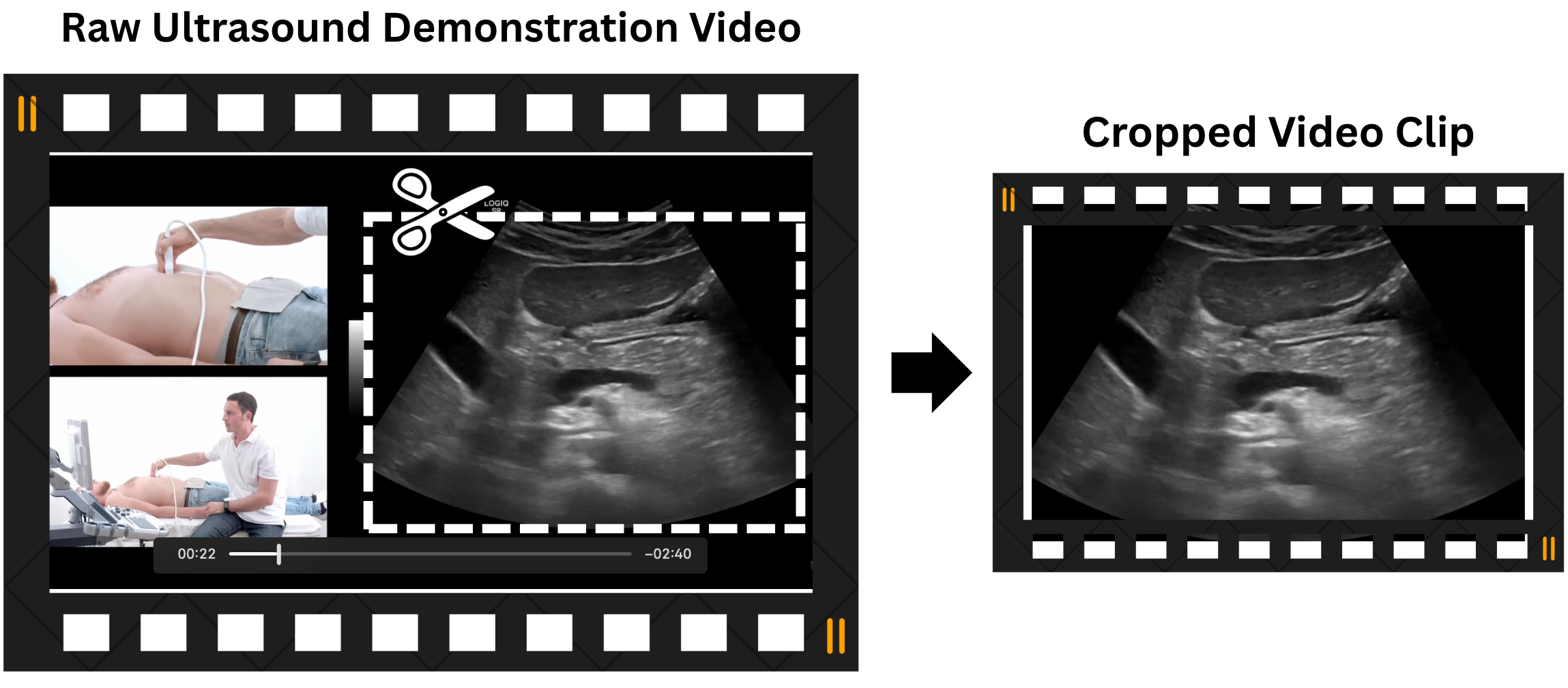}
  \caption{Example of video preprocessing. We crop the original video frame to retain only the ultrasound image stream, excluding surrounding content.}
  \label{fig:data_example}
\end{figure}

\paragraph{Data Source and Inclusion Criteria.} 
We curate ultrasound instructional videos from public sources (e.g., YouTube) based on four criteria:
\begin{itemize}
  \item \textbf{Procedure-rich Narration.} Clear voice-over explaining probe maneuvers, view targets, and troubleshooting rationale.
  \item \textbf{Continuous Scanning Segments.} Sustained scanning sequences rather than static frames.
  \item \textbf{Procedure Relevance.} Focus on acquisition workflow (probe movements, patient positioning, device adjustments, protocol progression) rather than diagnostic interpretation.
  \item \textbf{No On-screen Annotations.} Videos with instructional overlays (e.g., labeled anatomy) are excluded to avoid visual confounding.
\end{itemize}
Selected videos are manually trimmed to retain only active scanning segments and cropped to preserve only the ultrasound image stream, excluding interface elements (Fig.~\ref{fig:data_example}).

\paragraph{Transcription and Time Alignment.} 
We transcribe videos using WhisperX \citep{bain2022whisperx} to produce word-level timestamps. We apply light normalization using an LLM (GPT 5.2) to remove filler words and standardize terminology while preserving all scanning-relevant content: maneuver descriptions, view targets, and troubleshooting instructions.

% \begin{figure}[t]
%   \centering
%   \begin{promptbox}{Example Ground-Truth Event Log}
% \small
% \noindent
% \textbf{Event 1}\\
% \textbf{\textcolor{olive}{Time:}} 0.03--7.64 s\\
% \textbf{\textcolor{olive}{Action:}} Place transducer in the midline near the xiphoid; orient indicator cranially.\\
% \textbf{\textcolor{olive}{Intent:}} Identify the liver edge in sagittal view.

% \vspace{0.6em}

% \textbf{Event 2}\\
% \textbf{\textcolor{olive}{Time:}} 8.48--17.60 s\\
% \textbf{\textcolor{olive}{Action:}} Slide transducer to the patient's right while maintaining plane.\\
% \textbf{\textcolor{olive}{Intent:}} Bring the portal vein into view.
%   \end{promptbox}
%   \caption{Example of time-aligned procedural events derived from instructional narration and video.}
%   \label{box:gt_example}
% \end{figure}

\begin{figure*}[t]
  \centering
  \begin{minipage}[t]{0.48\textwidth}
    \begin{examplebox}[title=Ground-Truth Event Log]
\small
\textbf{Event 1} \hfill \textit{0.03--7.64\,s}\\[2pt]
\textcolor{slategray}{\textbf{Action:}} Place transducer in the midline near the xiphoid; orient indicator cranially.\\
\textcolor{slategray}{\textbf{Intent:}} Identify the liver edge in sagittal view.

\vspace{0.6em}

\textbf{Event 2} \hfill \textit{8.48--17.60\,s}\\[2pt]
\textcolor{slategray}{\textbf{Action:}} Slide transducer to the patient's right while maintaining plane.\\
\textcolor{slategray}{\textbf{Intent:}} Bring the portal vein into view.
    \end{examplebox}
  \end{minipage}%
  \hfill
  \begin{minipage}[t]{0.48\textwidth}
    \begin{examplebox}[title=Question-Answer Item]
\small
\textbf{Type:} Type~1 (Action--Goal) \\
\textbf{Format:} free-response \hfill \textit{Clip: 71.96--82.57\,s}\\[4pt]
\textcolor{slategray}{\textbf{Q:}} Based on the video clip, what sweep direction is used for the next segment and what is the imaging objective?\\[4pt]
\textcolor{slategray}{\textbf{A:}} The operator begins superiorly and sweeps inferiorly through the remaining segment, then returns to the starting position. The goal is to fully visualize the target region.
% \hfill \textit{Clip: 71.96--82.57\,s}
    \end{examplebox}
  \end{minipage}
  \caption{Examples from ReXSonoVQA. \textbf{Left:} Time-aligned procedural events derived from instructional narration. \textbf{Right:} A clip-grounded question-answer item targeting action-goal reasoning.}
  \label{fig:examples}
\end{figure*}

\paragraph{Ground Truth Event Log Construction.} 
Each video captures a single continuous ultrasound scan, which we represent as a standardized event log: a sequence of time-aligned procedural events (Fig.~\ref{fig:examples}). Each event is anchored to a temporal window (\textit{time\_start}, \textit{time\_end}) and captures a single coherent procedural unit with two components: ((i) \emph{action}, what the operator does (probe motion, device adjustments, patient repositioning), and (ii) \emph{intent}, the operator's goal (acquiring a target view, optimizing an acoustic window, confirming orientation, resolving ambiguity).
To construct the event log, we use an LLM (GPT 5.2) to convert the time-aligned transcript into structured, timestamped events. LLM-assisted extraction is effective here because instructional narration often describes procedural intent spanning multiple low-level actions (e.g., ``rotate to long axis,'' ``use the liver as an acoustic window,'' ``reduce sector width''), and the LLM distills these into structured representations directly usable for QA generation.

\subsection{Prompt Refinement and Quality Control Loop}
\label{sec:qc}

\paragraph{Initial Question Generation.} 
For each video, we generate candidate question-answer (QA) items by prompting an LLM (GPT 5.2) with a full time-segmented event log as input. The model produces a mixture of MCQs and free-response questions, each aligned to a clip window via \textit{time\_start} and \textit{time\_end}. Questions may derive from a single event or from adjacent event spans to form longer coherent procedural units (e.g., setup $\rightarrow$ maneuver $\rightarrow$ confirmation) (see Appendix~\ref{apd:prompts}, Fig.~\ref{box:prompt_a1.1} and~\ref{box:prompt_a1.2}).
Each QA item follows a standard format: (i) question (tagged as MCQ or free-response), (ii) answer (single correct option for MCQ; brief explanatory response for free-response), (iii) question type (Type 1/2/3), (iv) groundtruth (concise rephrasing fusing action and intent), and (v) \textit{time\_start}, \textit{time\_end} timestamps for clip extraction (Fig.~\ref{fig:examples}).
To control redundancy and maintain difficulty, the number of QA items generated scales with video duration and event density. Shorter videos yield proportionally fewer questions, reducing trivial or repetitive coverage of the same event.

% \begin{figure}[t]
%   \centering
%   \begin{promptbox}{Example QA}
% \small
% \noindent
% \textbf{Question Type:} Type~1 (Action--Goal Reasoning)\\
% \textbf{Format:} free-response\\ 
% \textbf{\textcolor{olive}{Question:}} Based on the video clip, what sweep direction is used for the next segment and what is the imaging objective of that pass?\\ 
% \textbf{\textcolor{olive}{Answer:}} The operator begins superiorly and sweeps inferiorly through the remaining segment, then returns to the starting position. The goal is to fully visualize and assess the target region across its extent.\\ 
% \textbf{\textcolor{olive}{Ground Truth:}} With a breath-hold, the operator starts superiorly and sweeps inferiorly through the target region (adjusting depth as needed), then sweeps back to complete the interrogation.\\ 
% \textbf{\textcolor{olive}{Clip Time:}} 71.96--82.57 s
%   \end{promptbox}
%   \caption{Example of a clip-grounded question-answer item targeting action-goal reasoning.}
%   \label{box:qa_example}
% \end{figure}

\begin{figure*}[t]
  \centering
  \includegraphics[width=\textwidth]{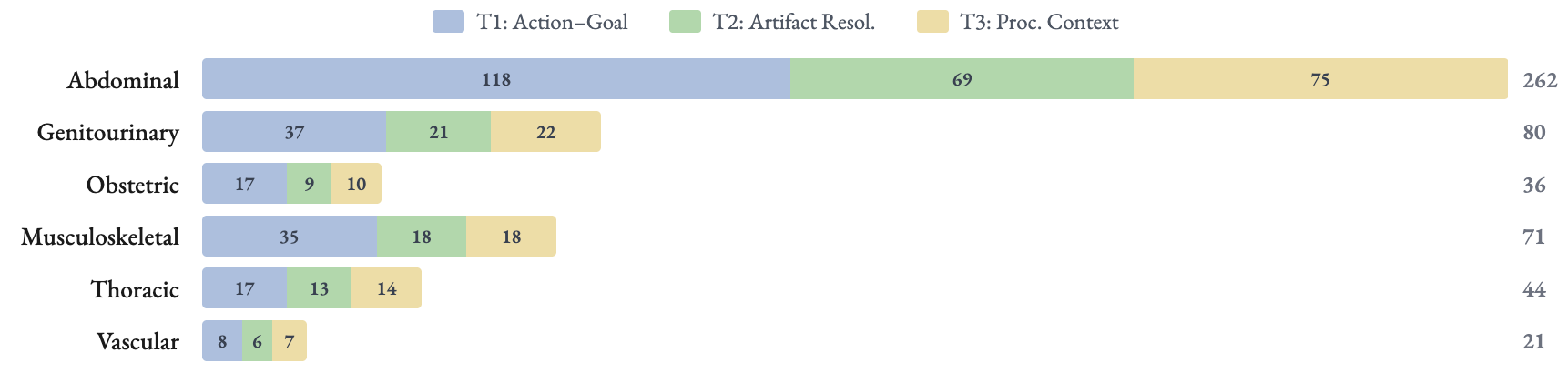}
  \caption{ReXSonoVQA dataset composition (514 items). Distribution of questions across clinical categories by task type.}
  \label{fig:data_compose}
\end{figure*}

\paragraph{Quality Control Loop.}
Unlike simple identification tasks, questions for procedural reasoning are more complex and can inadvertently contain answer cues in their phrasing, allowing models to guess without viewing the video. To ensure questions genuinely require video-based understanding, we apply an iterative quality control and prompt refinement loop.
\begin{itemize}
    \item \textbf{Free-Response QC.} We perform blind solvability screening by submitting each question to a VLM without the video. Items answered correctly are flagged as potentially solvable from generic clinical priors or answer-leaking phrasing. Flagged items are manually reviewed to distinguish questions answerable from wording alone from those answered correctly by chance.
    % genuinely trivial questions (answer inferable from wording) from those answered correctly by chance. 
    Trivial items are either (i) rewritten to remove leading cues, (ii) regenerated from the same event span with stricter prompt constraints, or (iii) removed from the benchmark.
    \item \textbf{MCQ QC.} We run the same text-only evaluation for MCQs, flagging items solved correctly without video. MCQs introduce an additional confound: the correct option may be identifiable through elimination if distractors are implausible or mismatched in abstraction level. During initial generation, we constrain the LLM to produce distractors that are (i) medically plausible within the ultrasound procedure context, (ii) at the same abstraction level as the correct answer (e.g., all maneuvers, all acquisition goals, or all troubleshooting actions), and (iii) compatible with the local anatomy or protocol rather than mixing unrelated organ workflows.

    To further improve distractor quality, we implement a post-processing step. For MCQs with weak incorrect options, we sample correct answers from other questions as replacement candidates, which are inherently more procedurally plausible. An LLM (GPT-5.2) then adapts these candidates to the local question context while keeping them incorrect (see Appendix~\ref{apd:prompts}, Fig.~\ref{box:prompt_a3}). We also shuffle option positions as the LLM disproportionately places correct answers.
\end{itemize}

\subsection{Benchmark Construction \& Evaluation}
\label{sec:stats}
The benchmark contains 514 time-localized clips/questions after QC, comprising 249 MCQs and 265 free-response items. The questions span a diverse set of ultrasound categories and scanning purposes, including abdominal, genitourinary, obstetric, musculoskeletal, thoracic, and vascular protocols (Fig.~\ref{fig:data_compose}). The current video collection contains 115.1 minutes of scanning footage after cropping. \textcolor{black}{Ground truth annotations derive from expert-narrated demonstration videos, and QA generation involved direct input from a board-certified clinician who participated in manual review of flagged items for medical accuracy.} Free-response predictions are scored using an LLM-as-a-judge approach, while MCQ performance is evaluated using standard accuracy against the ground-truth option.

% \begin{table}[!t]
%   \centering
%   % Move caption and label here for top-alignment
%   \caption{USVBench question counts by clinical category (after QC).}
%   \label{tab:dataset_stats}
%   \small
%   \setlength{\tabcolsep}{4pt}
%   \begin{tabular}{lrrr}
%     \toprule
%     \textbf{Category} & \textbf{MCQ} & \textbf{Free} & \textbf{Total} \\
%     \midrule
%     Abdominal         & 126 & 136 & 262 \\
%     Genitourinary     & 41  & 39  & 80  \\
%     Obstetric         & 18  & 18  & 36  \\
%     Musculoskeletal   & 33  & 38  & 71  \\
%     Thoracic          & 20  & 24  & 44  \\
%     Vascular          & 11  & 10  & 21  \\
%     \midrule
%     \textbf{Total}    & \textbf{249} & \textbf{265} & \textbf{514} \\
%     \bottomrule
%   \end{tabular}
% \end{table}

\section{Experiments}
\label{sec:experiments}

\subsection{Evaluation Method}
\label{sec:eval}

% We package the final benchmark as a per-scan, categorized JSON file. 
% Each file contains the full set of QA items with the schema defined earlier (question, answer, question type, fused ground-truth summary, and clip timestamps).

For evaluation, we perform zero-shot assessment \textcolor{black}{using Gemini 3 Pro, Qwen3.5-397B, LLaVA-Video-72B, and Seed 2.0 Pro} by submitting each QA item along with its corresponding video clip with audio track removed. MCQ items are scored using accuracy by comparing the selected option against the ground-truth option. Free-response items are graded using an LLM-as-a-judge protocol with a three-level rubric:
\begin{itemize}
  \item \textbf{Score 2 (Correct):} Both visual evidence and procedural reasoning/conclusion are correct.
  \item \textbf{Score 1 (Partially Correct):} Either the visual evidence is correct but the reasoning/conclusion is wrong, or the reasoning/conclusion is correct but the visual evidence is incorrect.
  \item \textbf{Score 0 (Incorrect):} Both visual evidence and reasoning/conclusion are incorrect.
\end{itemize}
This scheme captures procedure-relevant failure modes: misinterpreting visual evidence, incorrect causal reasoning, or both. \textcolor{black}{The judge receives the ground truth as context and is explicitly instructed to penalize hallucinations while applying criteria consistently across all questions and models.}

\subsection{Evaluation Settings}
\label{sec:blind}

To quantify the contribution of visual evidence and identify questions answerable without video, we evaluate the model under two matched input settings. Concretely, each QA item is tested under: (i) \textit{video-informed}, where the model receives the clip corresponding to $[\textit{time\_start}, \textit{time\_end}]$, and (ii) \textit{text-only}, where the model receives only the question (and MCQ options, if applicable) but no visual input. 
The text-only setting serves two purposes: it provides a lower-bound reference to quantify the marginal value of visual cues, and it acts as a quality diagnostic, items solvable without video indicate potential issues with question design (generic priors, answer-leaking phrasing, or MCQ option artifacts) and are flagged for revision during QC.

% The text-only setting serves as a quality diagnostic: items solvable without video indicate potential issues with question design—generic priors, answer-leaking phrasing, or MCQ option artifacts—rather than requiring true understanding from visual evidence.

% We use the blind setting in two ways. First, it provides a lower-bound reference against which video-informed performance can be compared to quantify the marginal value of visual cues. Second, it acts as a QA dataset quality indicator: blind-solvable questions are flagged for revision during our QC loop, and aggregate blind performance is used to assess whether the benchmark meaningfully requires video-based procedural reasoning.

% \subsection{Evaluation Results}
% \label{sec:results}

% We summarize Gemini 3 Pro performance for both MCQ accuracy and free response scoring (Fig.~\ref{fig:usvbench_results} and ~\ref{fig:usvbench_results_by_duration}), and compare against the blind (text-only) setting. To further characterize the effect of visual input, we report confusion matrices comparing video-informed and blind outcomes for MCQ and free response (Tables~\ref{tab:mcq_confusion} and~\ref{tab:fr_confusion}). We additionally include qualitative case studies (Fig.~\ref{fig:case_1} and~\ref{fig:case_2}, Appendix~\ref{apd:casestudies}) that highlight typical failure modes and serve as concrete examples of benchmark items.

\begin{figure*}[t]
  \centering
  \includegraphics[width=\textwidth]{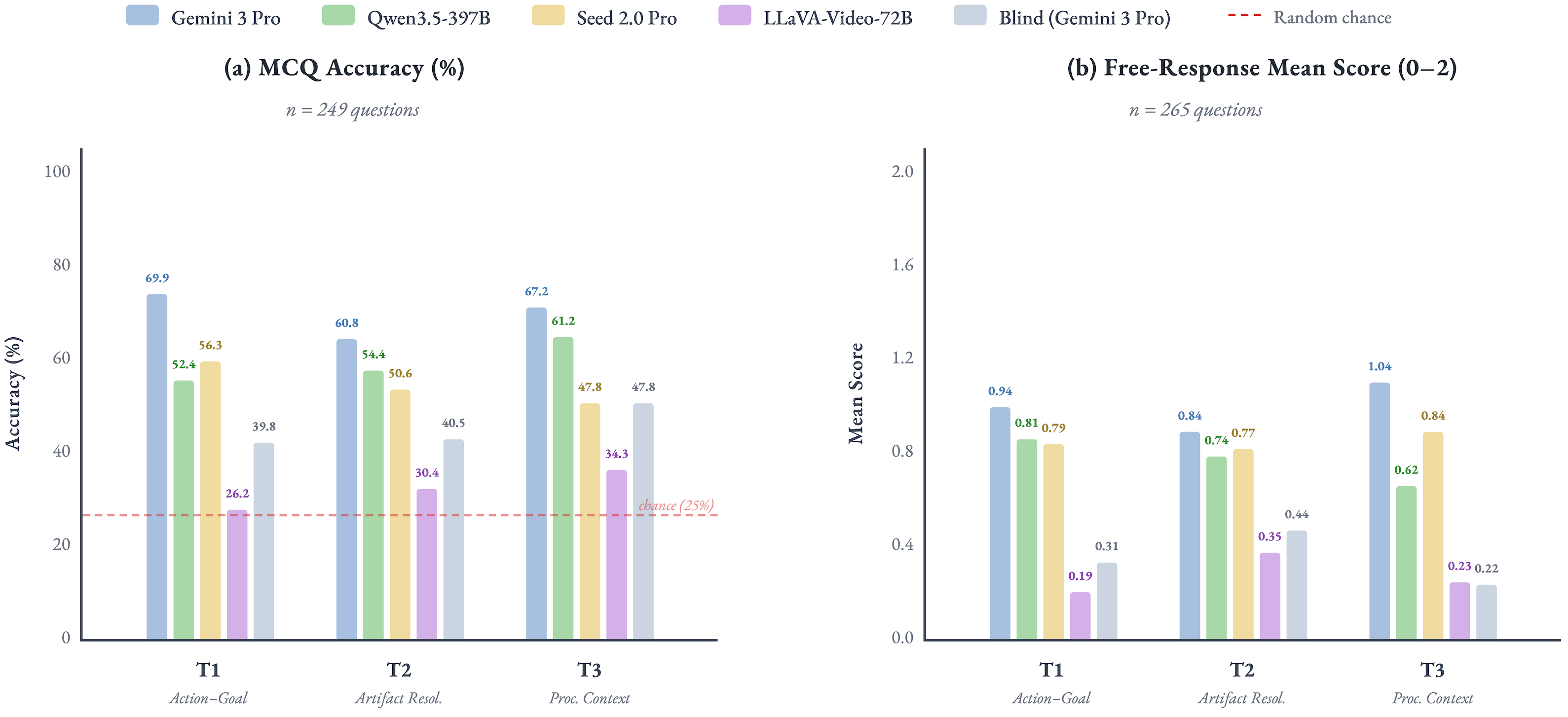}
  \caption{\textcolor{black}{Multi-model zero-shot MCQ accuracy (\%, left) and free-response mean score (0--2 rubric, right) by task type under paired evaluation settings. Dashed red line marks 25\% random chance for MCQ.}}
  \label{fig:results_by_type}
\end{figure*}
\begin{figure*}[t]
  \centering
+  \includegraphics[width=\textwidth]{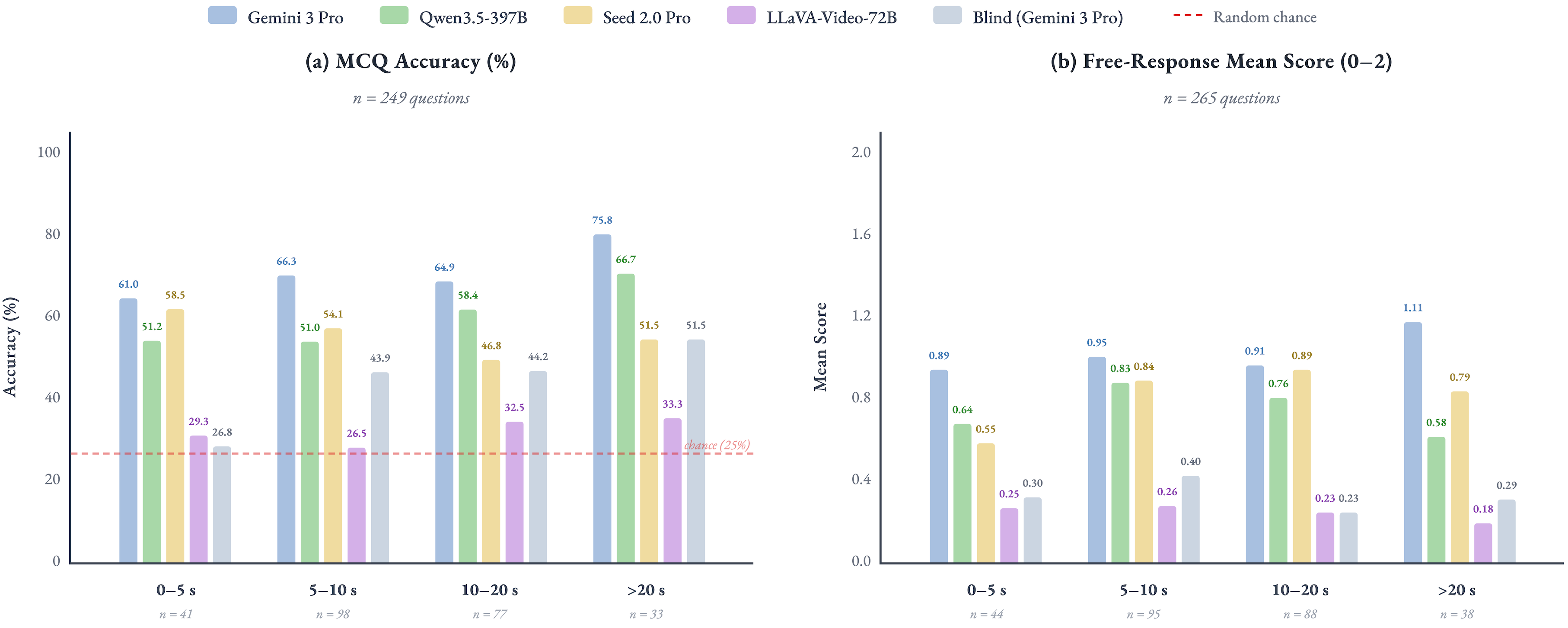}
  \caption{\textcolor{black}{Multi-model zero-shot MCQ accuracy (\%, left) and free-response mean score (0--2 rubric, right) by video duration under paired evaluation settings. Dashed red line marks 25\% random chance for MCQ.}}
  \label{fig:results_by_duration}
\end{figure*}

\section{Results}
\label{sec:results}
We evaluate \textcolor{black}{Gemini 3 Pro, Qwen3.5-397B, LLaVA-Video-72B, and Seed 2.0 Pro} under both video-informed and text-only (blind) settings to assess whether the benchmark meaningfully requires visual understanding. Results are summarized in Figs.~\ref{fig:results_by_type} and~\ref{fig:results_by_duration}, with cross-setting outcome tables in Appendix~\ref{apd:additional_results} (Tables~\ref{tab:mcq_confusion}--\ref{tab:fr_confusion_seed}).
% We summarize Gemini 3 Pro performance for both MCQ accuracy and free response scoring (Fig.~\ref{fig:usvbench_results} and ~\ref{fig:usvbench_results_by_duration}), and compare against the blind (text-only) setting. 
% To further characterize the effect of visual input, we report confusion matrices comparing video-informed and blind outcomes for MCQ and free response (Tables~\ref{tab:mcq_confusion} and~\ref{tab:fr_confusion}).
We additionally include qualitative case studies (Fig.~\ref{fig:case_1} and~\ref{fig:case_2}, Appendix~\ref{apd:casestudies}) that highlight typical failure modes and serve as concrete examples of benchmark items.

\paragraph{Task-wise Trends and the Role of Duration.}
\textcolor{black}{We present detailed analysis for Gemini 3 Pro, the best-performing model, followed by a multi-model comparison.} Across both MCQ and free-response evaluations, Gemini 3 Pro struggles most on Type~2 (Artifact Resolution \& Optimization) (Fig.~\ref{fig:results_by_type}). For MCQ, Type~1 reaches the highest accuracy (69.9\%), while Type~2 has the lowest (60.8\%), indicating that troubleshooting remains challenging even given the visual evidence is available. For free-response (0--2 rubric), Type~3 attains the highest mean score with video (\textcolor{black}{1.04}), whereas Type~2 again yields the lowest (\textcolor{black}{0.84}), reinforcing that resolving artifacts and ambiguity requires tighter causal reasoning beyond recognizing salient motion cues. 

Performance improves with clip duration across both formats: MCQ accuracy ranges from 61.0\% (0--5\,s) to 75.8\% ($>$20\,s), and free-response scores vary from \textcolor{black}{0.89} (0--5\,s) to \textcolor{black}{1.11} ($>$20\,s) (Fig.~\ref{fig:results_by_duration}). This pattern suggests longer clips provide richer temporal context that aids procedural reasoning. Both factors influence difficulty, with Type~2 tasks and short clips presenting the greatest challenges.

\textcolor{black}{\paragraph{Multi-Model Comparison.}
Figs.~\ref{fig:results_by_type} and~\ref{fig:results_by_duration} compare all four evaluated models. Gemini 3 Pro achieves the highest overall performance across both formats. Qwen3.5-397B and Seed 2.0 Pro show meaningful gains over the blind baseline, while LLaVA-Video-72B performs near or below blind levels, suggesting limited procedural video understanding. Across all models, Type~2 remains the most challenging on average on all tasks and question types, confirming that troubleshooting reasoning represents a fundamental challenge rather than a model-specific weakness.}

\paragraph{Visual Input Helps, but Not Uniformly.}
Visual input provides a clear overall benefit, but the margin depends on question type and format. For MCQ, overall accuracy improves from 42.2\% in the text-only setting to 66.3\% with video (+24.1 points), with the largest gain in Type~1 (+30.1) and the smallest in Type~3 (+19.4). For free-response, the overall mean score increases from \textcolor{black}{0.31} (text-only) to \textcolor{black}{0.95} (with video) (\textcolor{black}{+0.64}), with the weakest improvement for Type~2 (\textcolor{black}{+0.40}). 

Duration-stratified results reveal distinct patterns between formats. For MCQ, gains from video fluctuate across duration bins, while the text-only baseline consistently rises with clip length (from 26.8\% to 51.5\%). This pattern suggests that longer MCQ questions carry more contextual cues enabling partial answering without visual evidence. In contrast, free-response shows increasing gains from video with longer durations (from \textcolor{black}{+0.59} for 0--5\,s to \textcolor{black}{+0.82} for $>$20\,s), while the text-only baseline fluctuates rather than increasing monotonically. This indicates that free-response questions benefit more from extended temporal context and are less confounded by question length than MCQs, demonstrating stronger dependence on genuine visual understanding.

The cross-setting outcome tables (Tables~\ref{tab:mcq_confusion}--\ref{tab:fr_confusion}, Appendix~\ref{apd:additional_results}) reveal additional nuance for Gemini 3 Pro. While video context generally improves performance, it is not without noise: a small number of cases transition from correct to incorrect, suggesting that visual features can sometimes be misleading. More significantly, the bulk of previously incorrect responses remained unsolved even with video, indicating that the visual stream did not always provide the information needed to correct text-only guessing failures. \textcolor{black}{Specifically, in the MCQ cross-setting table, 68 items transition from incorrect (blind) to correct (with video), yielding a 47\% recovery rate (68/144); the 76 items remaining incorrect confirm substantial residual difficulty. The free-response cross-setting table reveals that video input selectively resolves perception gaps (64 items improve to full credit), while 98 items remain at 0 points and 22 items show correct perception but persistent reasoning errors, demonstrating that procedural understanding requires both visual perception and causal clinical reasoning.}

Taken together, these results confirm that the benchmark meaningfully relies on video, with free-response questions showing particularly strong dependence on visual evidence. However, troubleshooting questions (Type~2) remain the least reliably solved in terms of marginal gains and likely demand stronger causal grounding than current zero-shot VLMs provide.

\section{Limitations and Future Work}
\label{sec:limitations}

\paragraph{MCQ blind accuracy exceeds random guessing.} 
Blind accuracy substantially exceeds random chance (25\%), indicating exploitable patterns in question phrasing or option structure that models can leverage without viewing the video. This can arise from distractors that are not equally plausible, mismatched abstraction levels across options, or subtle linguistic cues favoring certain options. While our distractor polishing and blinding QC reduce this effect, further improvements include harder distractor generation, adversarial option auditing, and automated elimination-bias checks could push blind performance closer to chance. Notably, free-response questions exhibit much lower blind baselines and stronger dependence on visual evidence, suggesting they may be better suited for evaluating video-grounded procedural reasoning.

\paragraph{\textcolor{black}{Annotation scope and domain coverage.}}
Our ground-truth answers may under-specify clinically correct details that are visually plausible from the clip, which can complicate evaluation for open-ended questions where multiple specificity levels are acceptable. For example (Fig.~\ref{fig:RexSonoExm}), an answer key may refer to ``the tendon,'' while a model might correctly identify a more specific structure (e.g., within the carpal tunnel). We mitigate this through rubric-based grading with partial credit and plan to expand annotations with multiple acceptable reference answers. \textcolor{black}{Additionally, we deliberately selected instructional videos because expert narration provides natural supervision labels enabling scalable ground truth construction. Extending to in-hospital clinical footage is part of our proposed roadmap as clinical partnerships are established; the pipeline design allows integration of diverse video sources with minimal modification.}

\paragraph{\textcolor{black}{Model coverage.}}
\textcolor{black}{We evaluate Gemini 3 Pro, Qwen3.5-397B, LLaVA-Video-72B, and Seed 2.0 Pro, all supporting native video input (Figs.~\ref{fig:results_by_type} and~\ref{fig:results_by_duration}). Existing medical VLMs rely solely on image-to-language architectures without video capability,} and other leading commercial VLMs (GPT-5.2, Claude Opus 4.5) lack native video support. As video-language models with medical-specific pretraining emerge, future work will extend evaluation to more models.

\section{Conclusion}
\label{sec:conclusion}

We introduce ReXSonoVQA, a procedure-centric ultrasound video QA benchmark evaluating VLMs' understanding of ultrasound scanning technique. Built from instructional videos with time-aligned event logs, it comprises 514 clip-grounded questions (249 MCQ, 265 free-response) across six clinical categories and three procedural competencies: Action-Goal Reasoning, Artifact Resolution \& Optimization, and Procedure Context \& Planning. An iterative quality control loop ensures questions require genuine video-based reasoning through text-only screening and distractor refinement.

Zero-shot evaluation \textcolor{black}{across four models (Gemini 3 Pro, Qwen3.5-397B, LLaVA-Video-72B, and Seed 2.0 Pro)} reveals that while VLMs can extract procedural information from video, substantial gaps remain. Type 1 (Action-Goal) achieves highest MCQ performance while Type 3 (Procedure Context) excels in free-response, but Type 2 (Artifact Resolution) shows minimal improvement over text-only baseline, exposing limitations in causal troubleshooting reasoning. Performance improves with clip duration, with free-response questions demonstrating stronger dependence on visual evidence. However, most text-only failures remain unsolved even with video, indicating fundamental procedural understanding gaps.

% Our analysis also reveals two key limitations. First, MCQ text-only accuracy substantially exceeds random chance, especially for longer clips where the text-only baseline performance rises consistently. This indicates that MCQ questions contain exploitable linguistic patterns, requiring better distractor generation methods and rigorous testing. In contrast, free-response questions demonstrate less confounding from question length, with fluctuating text-only baselines and increasing gains from video, suggesting they more reliably measure genuine visual procedural understanding. Second, ground-truth annotations may under-specify acceptable responses. When model predictions contain more anatomical or procedural detail than our reference answers, we cannot reliably determine whether the additional information is accurate or hallucinated.

ReXSonoVQA establishes the first evaluation framework for video-based procedural ultrasound understanding, providing a foundation for developing VLM perception systems critical to ultrasound training, real-time clinical guidance, and robotic automation.

\bibliography{chil-sample}

\appendix

\setcounter{figure}{0} 

\renewcommand{\thefigure}{A\arabic{figure}}
\renewcommand{\theHfigure}{A\arabic{figure}}

\setcounter{table}{0}

 \renewcommand{\thetable}{A\arabic{table}}
\renewcommand{\theHtable}{A\arabic{table}}
\section{Prompts}
\label{apd:prompts}
See Fig~\ref{box:prompt_a1.1},~\ref{box:prompt_a1.2},~\ref{box:prompt_a2},~\ref{box:prompt_a3},~\ref{box:prompt_a4}
% ---- Prompt A1 ----
\begin{figure*}[t]
  \centering
  \begin{examplebox}[title={Prompt A1: QA generation}]
\small
\begin{lstlisting}
You are generating Q&A items for a benchmark that evaluates a Vision-Language Model (VLM) on ULTRASOUND DEMONSTRATION VIDEOS.

CRITICAL EVAL SETUP (what the VLM will see at test time):
- Video ONLY (muted). No audio.
- Therefore, every question must be answerable from VISUAL cues only:
  - probe motion (slide/sweep/rock/rotate/tilt, directionality, marker orientation when visible),
  - patient positioning/motion if visually observable,
  - ultrasound image dynamics (structures entering/leaving frame, plane changes, motion, shadowing),
  - mode changes that are visually obvious (e.g., Color Doppler overlay appears/disappears),
  - imaging parameter changes if visible in the image effect (e.g., depth scale changes, focal zone marker moves, sector narrows).

GROUND TRUTH YOU WILL RECEIVE:
- A sequence of time-stamped events extracted from transcript, each with:
  - action (what operator did)
  - interpretation (why / what they were looking for)
These are NOT available to the VLM at test time; they are for YOU to craft correct QA.

THE MOST CRITICAL QUESTION CONSTRUCTION RULE:
Each question must be answerable only by watching the video. Do not include visual clues, descriptions, or contextual hints that would allow someone to answer or guess correctly without viewing the video. 
Do not contain specific visual details, structures, context, artifacts, or outcomes in the question itself.

BAD (includes visual clues): "When the operator sweeps inferiorly and identifies the bifurcation, what anatomical landmark is being visualized?"
GOOD (requires watching): "Based on the clip, what acquisition goal is being pursued during the continuous inferior sweep, and what key anatomic endpoint indicates success?"

BAD (describes what's visible): "As bowel gas obscures the vessel view and the operator applies firm pressure..."
GOOD (asks about action/goal): "The clip shows loss/obscuration of the target vessel. What optimization maneuver is performed, and what problem is it addressing to restore the view?"

Always phrase questions as:
- "Based on the video/clip, what..."
- "What [action/strategy/technique] is shown, and what is the goal/objective?"
- "This clip shows [general/non-specific observation]. What is being done and why?"

You must generate items in exactly 3 clinically-meaningful question types:

1) Type1_ActionGoalReasoning
   - Tests: action reasoning + goal inference.
   - Ask: what maneuver is being performed AND what imaging goal / target view it serves.
   - Phrase questions naturally and adaptively based on the content.
   - Do NOT describe the specific anatomical structures or visual details in the question.

2) Type2_ArtifactResolutionOptimization
   - Tests: overcoming artifacts or ambiguity + optimization/disambiguation logic.
   - Ask: what (probe maneuver, patient management, or knobology) has changed AND why it resolves an artifact or ambiguity / improves image quality.
   - IMPORTANT: Do not explicitly describe the artifact or ambiguity in the question.
   - Phrase questions naturally - you can reference general observations like "loss of view" or "poor quality" without describing specific details.

\end{lstlisting}
  \end{examplebox}
  \caption{Prompt template used for QA generation.}
  \label{box:prompt_a1.1}
\end{figure*}

% ---- Prompt A2 ----
\begin{figure*}[t]
  \centering
  \begin{examplebox}[title={Prompt A1(continued): QA generation}]
\small
\begin{lstlisting}
3) Type3_ProcedureContextPlanning
   - Tests: overall context understanding + next-step planning.
   - Ask: what phase/step the operator is in AND what the broader workflow objective or next logical step is.
   - Usually use TWO or more ADJACENT EVENTS to create sufficient context.
   - Vary your phrasing - ask about exam phases, workflow transitions, procedural objectives, or strategies as appropriate.
   - Do not describe the clip's content or sequence and do not name specific anatomy or maneuvers in the question itself.

QUESTION FORMAT MIX:
- Produce a mix of MCQ and free response.
- For MCQ: 
  * Put exactly 4 options (A-D) inside the "question" string. Ensure only ONE is correct; distractors must be plausible.
  * Keep options generic and hypothesis-based when possible
  * Do NOT make options that give away visual details that should only be known from watching the video
  * IMPORTANT: The "answer" field must contain ONLY the letter of the correct option (e.g., "A" or "B" or "C" or "D")
- For Free response: 
  * The "question" must be answerable concisely; the "answer" should be 1-3 sentences.

QUESTION PHRASING PRINCIPLES:
- Vary your question style naturally based on the content and question type
- Ask WHAT is being done and WHY, without describing what is visible

TIME GROUNDING:
- Each item must include time_start and time_end aligned to the event(s) used:
  - Single event: use its exact start/end
  - Two or more adjacent events: time_start = min(starts), time_end = max(ends)

GROUNDTRUTH FIELD:
- Rephrase the action + interpretation into ONE integrated entry (1 to 3 sentences) that preserves all key details.
- Do not introduce details not supported by action/interpretation.

AVOID INVALID ITEMS:
- Do NOT reference audio, narration, or transcript.
- Do NOT ask for reading on-screen text or exact numeric readouts (e.g., "104 mm") unless it is guaranteed without OCR (assume it is not).
- If an action is a spoken instruction (e.g., "hold breath"), convert it into an EFFECT-based question that could be inferred visually (e.g., reduced respiratory motion / improved acoustic window), not "what instruction was said".

Return JSON only, matching the schema provided.
\end{lstlisting}
  \end{examplebox}
  \caption{Prompt template used for QA generation (continued).}
  \label{box:prompt_a1.2}
\end{figure*}

% ---- Prompt A3 ----
\begin{figure*}[t]
  \centering
  \begin{examplebox}[title={Prompt A2: MCQ distractor refinement}]
\small
\begin{lstlisting}
You are a medical MCQ editor. Improve distractors so they are clinically plausible and not trivially wrong.

Rules:
- Do NOT change the question stem.
- Do NOT change the correct option text.
- Provide only the requested distractors.
- Distractors must be medically plausible and consistent with the exam context, but clearly incorrect.
- Avoid nonsensical or unrelated anatomy/procedures.
- Keep each option concise (1 sentence max).
- Do not include letters like 'A.' in the option text; return raw option text only.
""".strip()

	user_payload = {
		"stem": stem,
		"correct_option": correct_text,
		"distractor_letters": distractor_letters,
		"exemplar_options": exemplar_options,
	}
\end{lstlisting}
  \end{examplebox}
  \caption{Prompt template used for MCQ distractor refinement.}
  \label{box:prompt_a2}
\end{figure*}

\begin{figure*}[t]
  \centering
  \begin{examplebox}[title={Prompt A3: Inference}]
\small
\begin{lstlisting}
You are an expert ultrasound clinician evaluating video clips from ultrasound examinations.

You will be shown a video clip and asked a question about it. The video contains ultrasound examination footage with the audio muted. You must answer based ONLY on what you observe visually in the video.

Key Guidelines:
- Watch the video carefully, observing probe movements, image changes, and anatomical structures
- Answer based solely on visual evidence from the video
- For multiple choice questions (MCQ): Start your response with "Answer: X" where X is the correct letter (A, B, C, or D), then provide a brief explanation.
- For free response questions: Provide a concise, accurate answer in 1-3 sentences
- If you cannot determine the answer from the video, state that clearly

Your response should be direct and factual, based on what you observe in the video clip.
\end{lstlisting}
  \end{examplebox}
  \caption{Prompt template used for model inference.}
  \label{box:prompt_a3}
\end{figure*}

\begin{figure*}[t]
  \centering
  \begin{examplebox}[title={Prompt A4: LLM-as-a-judge for Free-Responses}]
\small
\begin{lstlisting}
STRICTNESS OVERRIDE:
- Penalize guessing or vague organ references.
- If the ground truth names a specific organ/structure/feature and the prediction only gives a generic organ reference (e.g., "an organ" or a different unspecified organ), treat this as incorrect visual evidence.
- If the prediction is uncertain or overly general (e.g., "likely an organ" without identification), do NOT award full credit. At most score=1, and classify the error as wrong_visual_evidence.
- Non-identification of the required organ/feature should be categorized as visual evidence error.
""".strip()

        system_prompt = """You are an impartial medical expert judge evaluating the quality of an AI assistant's response to a clinical ultrasound question.

    Compare the AI's Prediction against the Ground Truth Answer.

    Scoring Criteria:
    2 - Correct: The conclusion/answer matches the ground truth AND the cited visual evidence aligns with the ground truth. Minor phrasing differences are acceptable.
    1 - Partially Correct: Only one of the two is correct (either the conclusion is correct but the visual evidence is wrong, OR the visual evidence is correct but the conclusion is wrong).
    0 - Incorrect: The prediction is wrong, irrelevant, hallucinated, or contradicts the ground truth.

    If score = 1, classify the mistake type:
    - wrong_visual_evidence: conclusion is correct, but evidence/visual justification is incorrect or mismatched.
    - wrong_conclusion: evidence/visual cues are correct, but the conclusion/answer is wrong.

    If score = 0, classify the mistake type:
    - wrong_visual_evidence: evidence/visual justification is incorrect.
    - wrong_conclusion: conclusion/answer is incorrect.
    - both_fail: both evidence and conclusion are incorrect.

    Output exactly and ONLY valid JSON in this format:
    {"score": 0, "explanation": "Brief reasoning", "error_type": "none"}
\end{lstlisting}
  \end{examplebox}
  \caption{Prompt template used for LLM-as-a-judge.}
  \label{box:prompt_a4}
\end{figure*}

\section{Case Studies}
\label{apd:casestudies}
\begin{figure*}[h]
  \centering
  \includegraphics[width=0.85\textwidth]{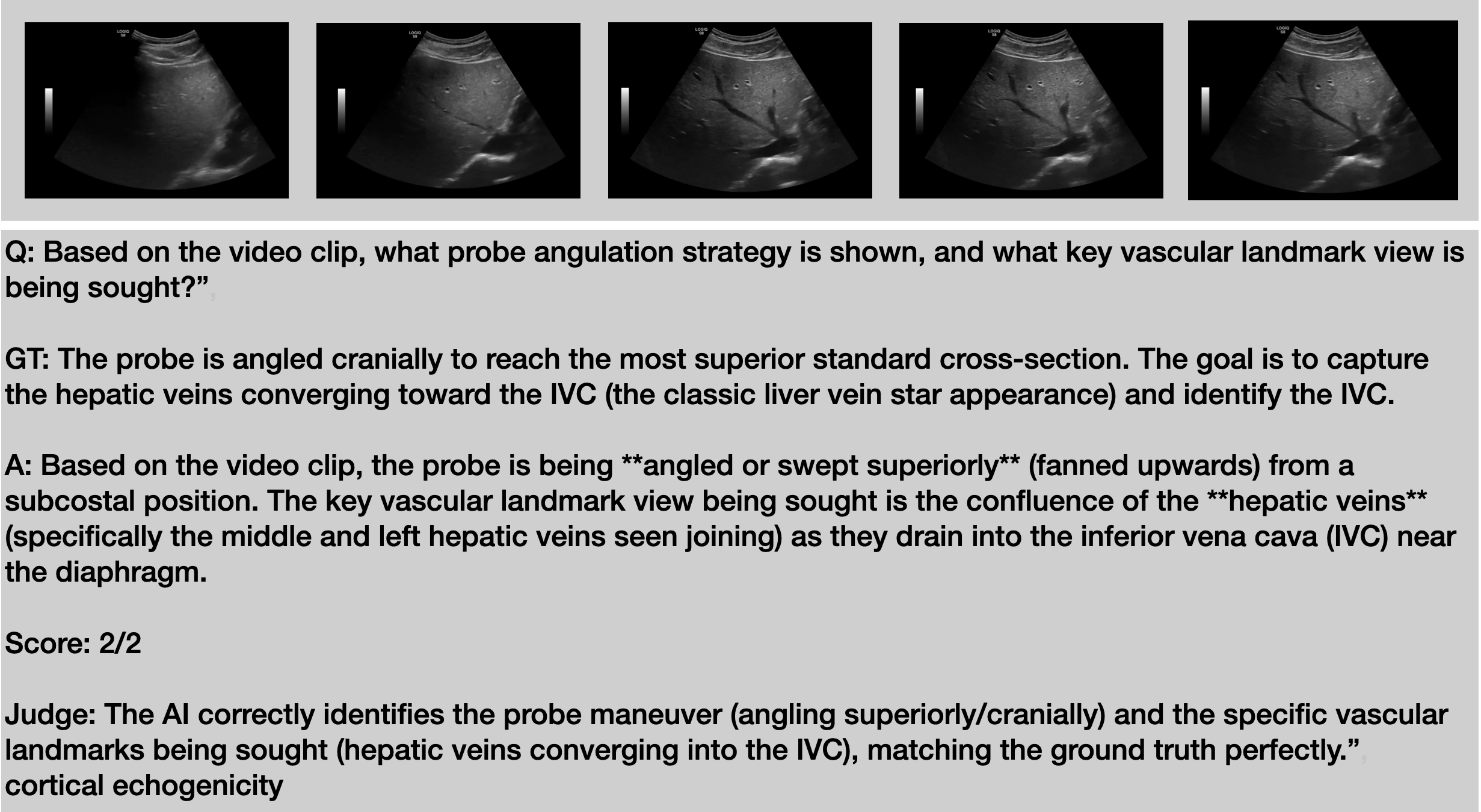}
  \caption{\textbf{Case Study 1 (Free-Response, Type~1: Action–Goal Reasoning).} Gemini 3 Pro correctly identifies the probe maneuver and acquisition target.}
  \label{fig:case_1}
\end{figure*}

\begin{figure*}[h]
  \centering
  \includegraphics[width=0.80\textwidth]{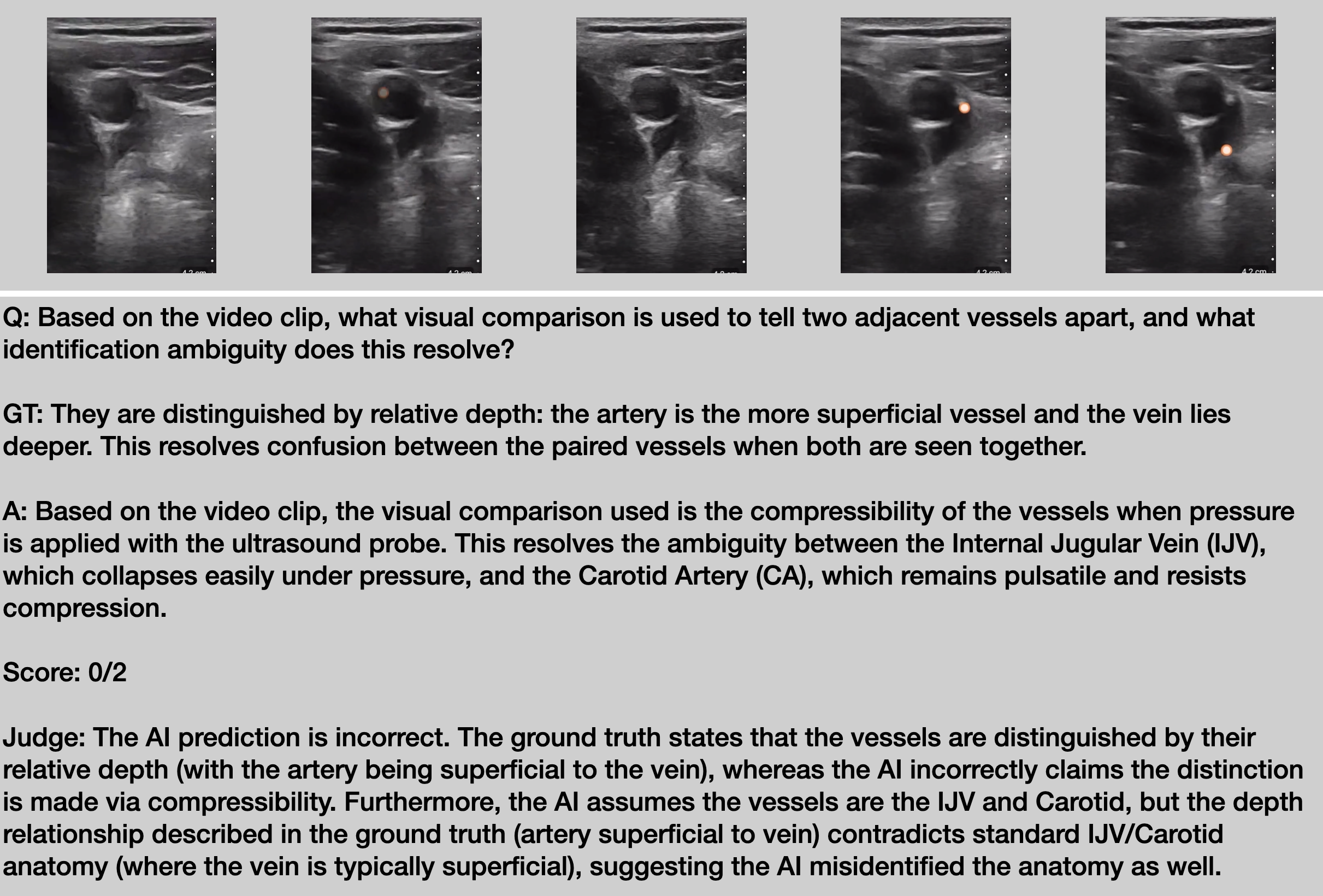}
  \caption{\textbf{Case Study 2 (Free-Response, Type~2: Artifact Resolution \& Optimization).} Gemini 3 Pro hallucinates a compression maneuver instead of the actual optimization action.}
  \label{fig:case_b2}
\end{figure*}

\begin{figure*}[h]
  \centering
  \includegraphics[width=0.80\textwidth]{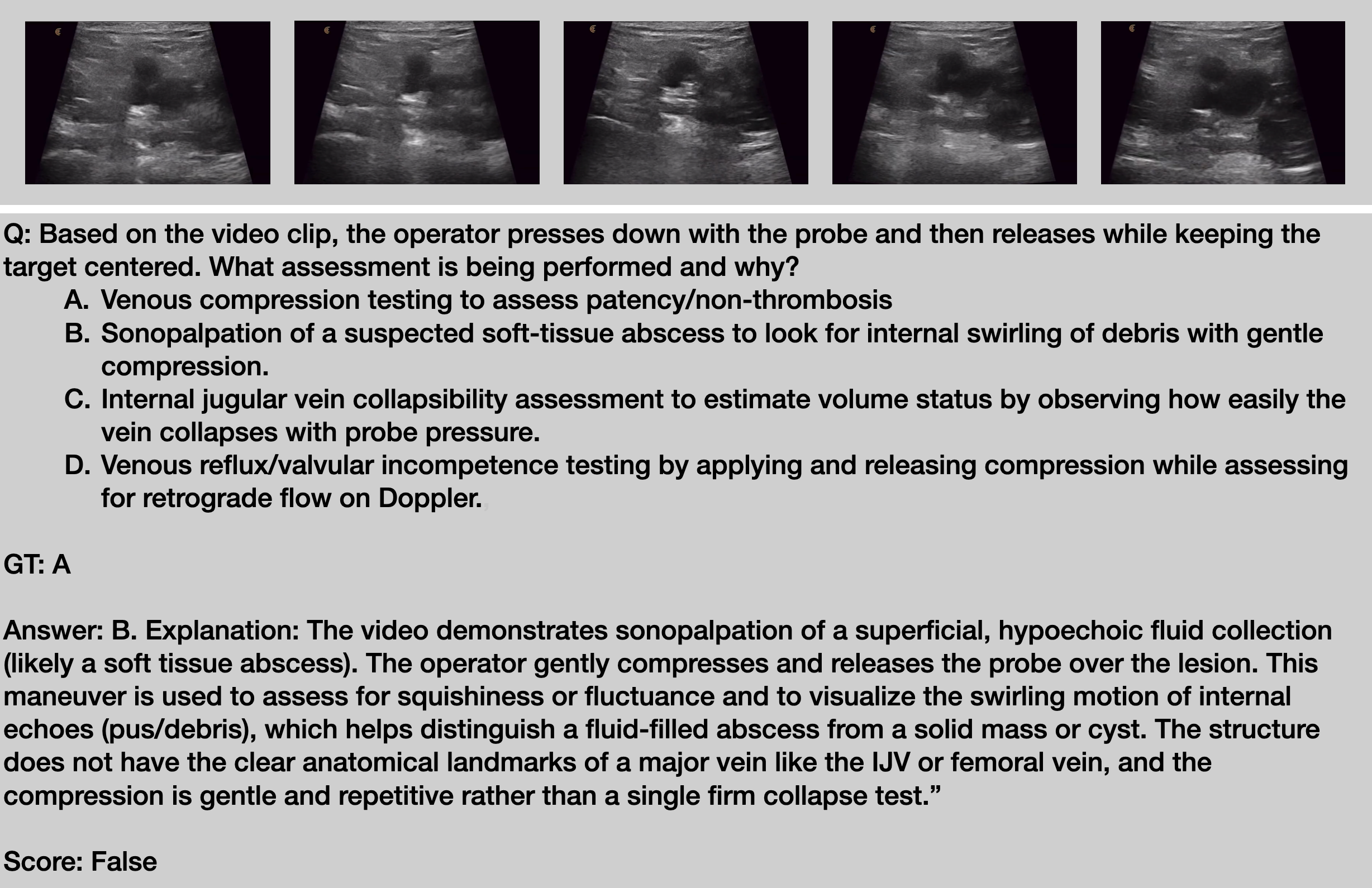}
  \caption{\textbf{Case Study 3 (MCQ, Type~1: Action–Goal Reasoning).} Gemini 3 Pro misinterprets compression behavior due to incorrect anatomy identification, leading to the wrong action–goal selection.}
  \label{fig:case_b3}
\end{figure*}

\begin{figure*}[h]
  \centering
  \includegraphics[width=0.80\textwidth]{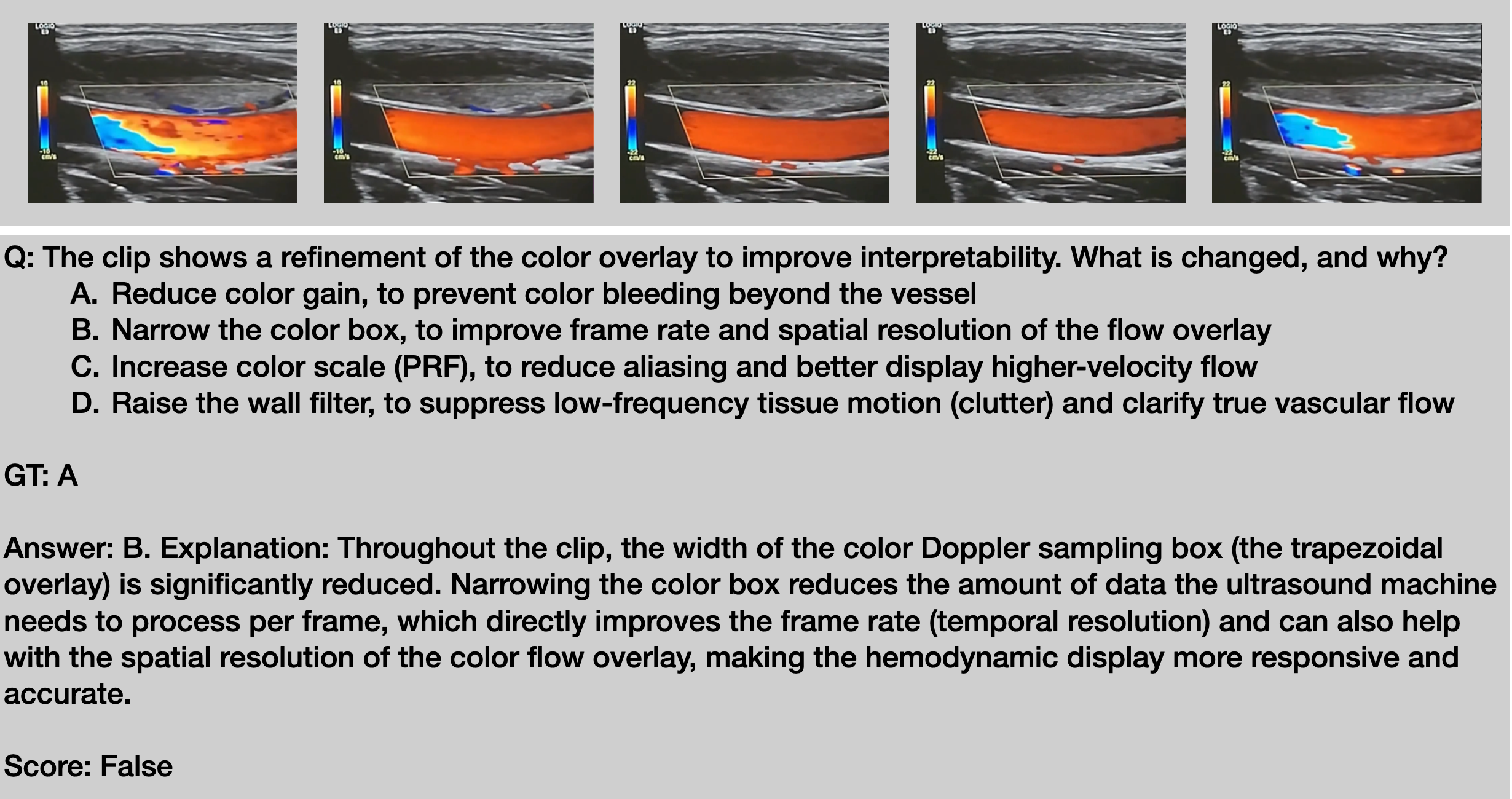}
  \caption{\textbf{Case Study 4 (MCQ, Type~2: Artifact Resolution \& Optimization).} Gemini 3 Pro fails to identify the color bleeding artifact and the corresponding optimization cue.}
  \label{fig:case_2}
\end{figure*}

\begin{figure*}[h]
  \centering
  \includegraphics[width=0.80\textwidth]{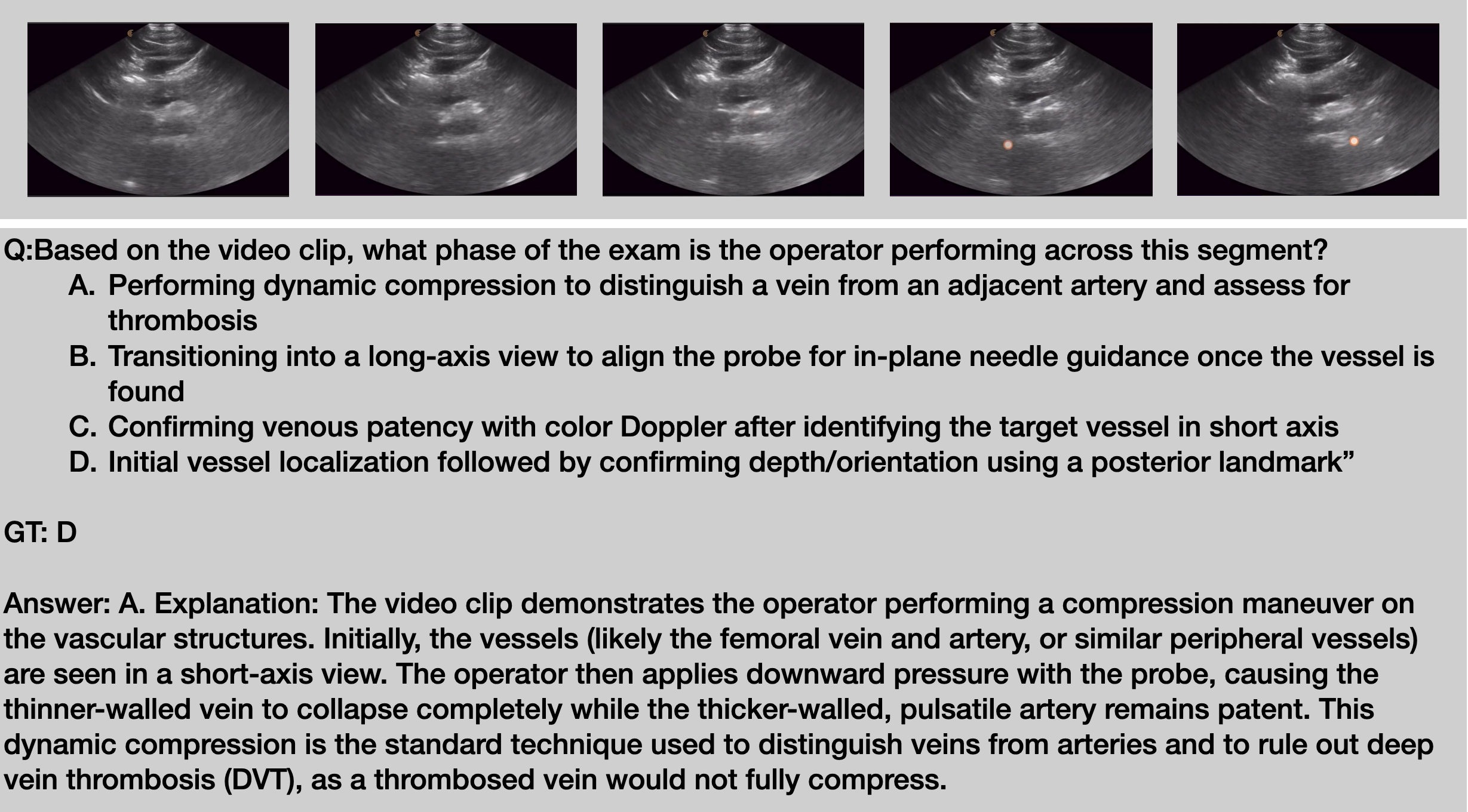}
  \caption{\textbf{Case Study 5: (MCQ, Type~3: Procedure Context \& Planning)} fails to recognize the vertebral acoustic shadow as the key landmark for protocol context.}
  \label{fig:case_b4}
\end{figure*}

See
Fig~\ref{fig:case_1},~\ref{fig:case_b2},~\ref{fig:case_b3},~\ref{fig:case_2},~\ref{fig:case_b4}
% -------- Case study image 1 --------

\begin{figure*}[h]
  \centering
  \includegraphics[width=0.85\textwidth]{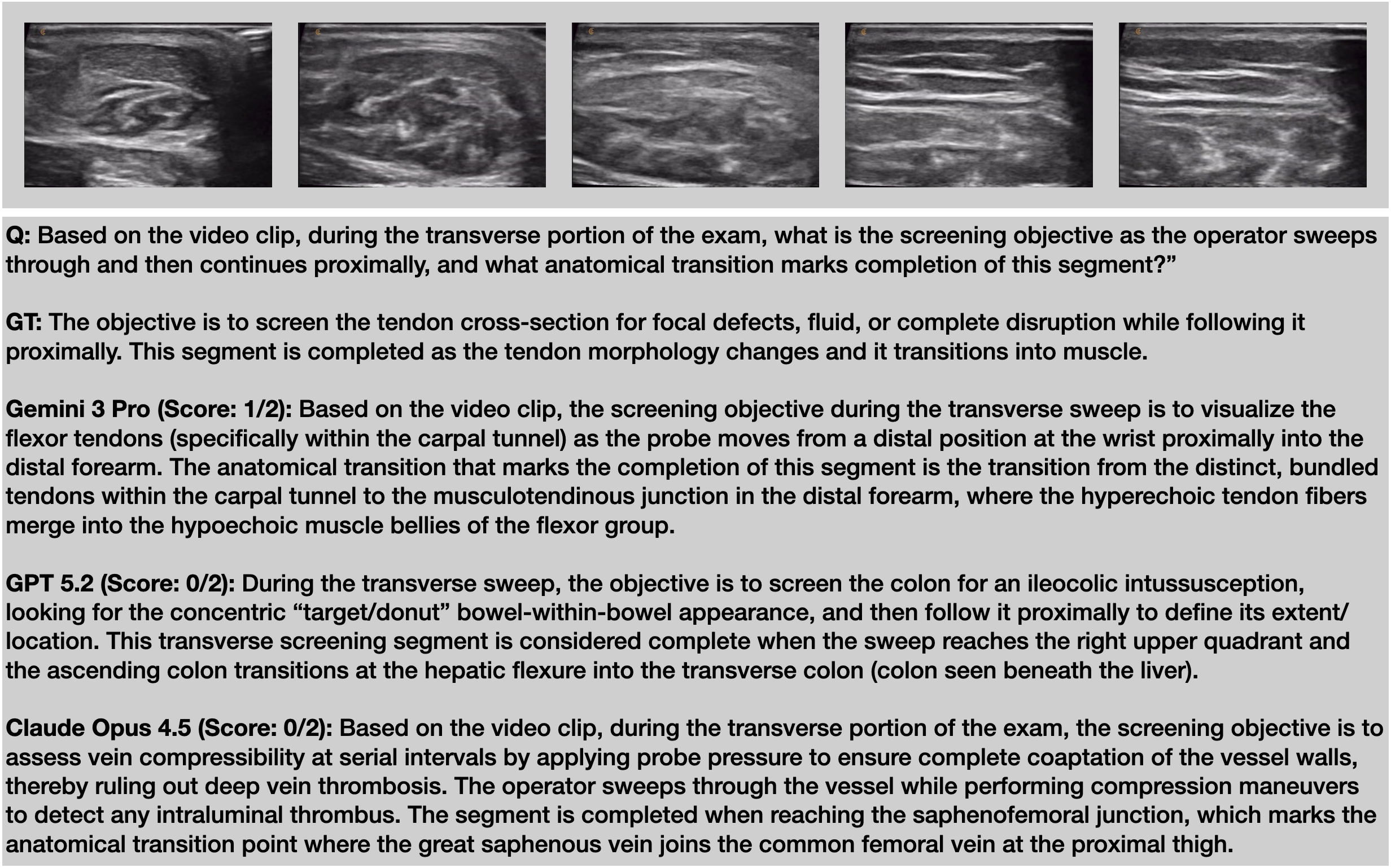}
  \caption{\textbf{Exploratory evaluation.} Demonstrated why GPT-5.2 and Claude Opus 4.5 were excluded from our benchmark's baseline. Without native video support, both models hallucinate completely unrelated anatomical structures (colon/intussusception and vein/thrombosis) when provided with sequential images, scoring 0/2. In contrast, Gemini 3 Pro with native video understanding correctly identifies the anatomical transition, achieving 1/2. This stark difference motivated our focus on models with native video support.}
  \label{fig:explore}
\end{figure*}

% \begin{figure*}[h]
%   \centering
%   \includegraphics[width=0.80\textwidth]{images/case1.png}
%   \caption{Case Study 3: Free-response (Type 1) fails to infer the intended next-step/action goal despite observing the ongoing scan context.}
%   \label{fig:case_b1}
% \end{figure*}

% -------- Case study image 2 --------

% -------- Case study image 3 --------

\section{Cross-Setting Outcome Tables}
\label{apd:additional_results}
Tables~\ref{tab:mcq_confusion}--\ref{tab:fr_confusion_seed} report the cross-tabulation of video-informed vs.\ text-only (blind) outcomes for all four evaluated models, separately for MCQ and free-response formats. Each cell counts the number of items falling into a given (blind outcome, video-informed outcome) pair, providing a fine-grained view of how visual input changes model predictions.

%%% ─── Gemini 3 Pro ───────────────────────────────────────────────────

\begin{table}[h]
\centering
\caption{Gemini 3 Pro: MCQ cross-setting outcome table ($n=249$). Rows = blind (text-only) outcome; columns = video-informed outcome.}
\label{tab:mcq_confusion}
\resizebox{\columnwidth}{!}{%
\small
\setlength{\tabcolsep}{5pt}
\begin{tabular}{@{}lrrr@{}}
\toprule
 & \textbf{Vid.\ Correct} & \textbf{Vid.\ Incorrect} & \textbf{Total} \\
\midrule
\textbf{Blind Correct}   & 97  & 8   & 105 \\
\textbf{Blind Incorrect}  & 68  & 76  & 144 \\
\midrule
\textbf{Total}            & 165 & 84  & 249 \\
\bottomrule
\end{tabular}%
}
\end{table}

\begin{table}[h]
\centering
\caption{Gemini 3 Pro: Free-response cross-setting outcome table ($n=265$). Rows = blind outcome; columns = video-informed outcome. W.C.\ = wrong conclusion; W.V.\ = wrong visual evidence.}
\label{tab:fr_confusion}
\resizebox{\columnwidth}{!}{%
\small
\setlength{\tabcolsep}{3pt}
\begin{tabular}{@{}lrrrrr@{}}
\toprule
 & \textbf{2\,pt} & \textbf{1\,pt\,(W.C.)} & \textbf{1\,pt\,(W.V.)} & \textbf{0\,pt} & \textbf{Total} \\
\midrule
\textbf{2\,pt}          & 15 & 2  & 1  & 1   & 19  \\
\textbf{1\,pt (W.C.)}   & 7  & 4  & 2  & 8   & 21  \\
\textbf{1\,pt (W.V.)}   & 13 & 1  & 3  & 6   & 23  \\
\textbf{0\,pt}          & 64 & 22 & 18 & 98  & 202 \\
\midrule
\textbf{Total}          & 99 & 29 & 24 & 113 & 265 \\
\bottomrule
\end{tabular}%
}
\end{table}

%%% ─── LLaVA-Video-72B ───────────────────────────────────────────────

\begin{table}[h]
\centering
\caption{LLaVA-Video-72B: MCQ cross-setting outcome table ($n=249$).}
\label{tab:mcq_confusion_llava}
\resizebox{\columnwidth}{!}{%
\small
\setlength{\tabcolsep}{5pt}
\begin{tabular}{@{}lrrr@{}}
\toprule
 & \textbf{Vid.\ Correct} & \textbf{Vid.\ Incorrect} & \textbf{Total} \\
\midrule
\textbf{Blind Correct}   & 45  & 60  & 105 \\
\textbf{Blind Incorrect}  & 29  & 115 & 144 \\
\midrule
\textbf{Total}            & 74  & 175 & 249 \\
\bottomrule
\end{tabular}%
}
\end{table}

\begin{table}[h]
\centering
\caption{LLaVA-Video-72B: Free-response cross-setting outcome table ($n=265$).}
\label{tab:fr_confusion_llava}
\resizebox{\columnwidth}{!}{%
\small
\setlength{\tabcolsep}{3pt}
\begin{tabular}{@{}lrrrrr@{}}
\toprule
 & \textbf{2\,pt} & \textbf{1\,pt\,(W.C.)} & \textbf{1\,pt\,(W.V.)} & \textbf{0\,pt} & \textbf{Total} \\
\midrule
\textbf{2\,pt}          & 0  & 2  & 6  & 11  & 19  \\
\textbf{1\,pt (W.C.)}   & 1  & 1  & 3  & 16  & 21  \\
\textbf{1\,pt (W.V.)}   & 4  & 2  & 3  & 14  & 23  \\
\textbf{0\,pt}          & 2  & 16 & 16 & 168 & 202 \\
\midrule
\textbf{Total}          & 7  & 21 & 28 & 209 & 265 \\
\bottomrule
\end{tabular}%
}
\end{table}

%%% ─── Qwen3.5-397B ──────────────────────────────────────────────────

\begin{table}[h]
\centering
\caption{Qwen3.5-397B: MCQ cross-setting outcome table ($n=249$).}
\label{tab:mcq_confusion_qwen}
\resizebox{\columnwidth}{!}{%
\small
\setlength{\tabcolsep}{5pt}
\begin{tabular}{@{}lrrr@{}}
\toprule
 & \textbf{Vid.\ Correct} & \textbf{Vid.\ Incorrect} & \textbf{Total} \\
\midrule
\textbf{Blind Correct}   & 66  & 39  & 105 \\
\textbf{Blind Incorrect}  & 72  & 72  & 144 \\
\midrule
\textbf{Total}            & 138 & 111 & 249 \\
\bottomrule
\end{tabular}%
}
\end{table}

\begin{table}[h]
\centering
\caption{Qwen3.5-397B: Free-response cross-setting outcome table ($n=265$).}
\label{tab:fr_confusion_qwen}
\resizebox{\columnwidth}{!}{%
\small
\setlength{\tabcolsep}{3pt}
\begin{tabular}{@{}lrrrrr@{}}
\toprule
 & \textbf{2\,pt} & \textbf{1\,pt\,(W.C.)} & \textbf{1\,pt\,(W.V.)} & \textbf{0\,pt} & \textbf{Total} \\
\midrule
\textbf{2\,pt}          & 13 & 1  & 1  & 4   & 19  \\
\textbf{1\,pt (W.C.)}   & 7  & 0  & 3  & 11  & 21  \\
\textbf{1\,pt (W.V.)}   & 9  & 2  & 5  & 7   & 23  \\
\textbf{0\,pt}          & 45 & 13 & 23 & 121 & 202 \\
\midrule
\textbf{Total}          & 74 & 16 & 32 & 143 & 265 \\
\bottomrule
\end{tabular}%
}
\end{table}

%%% ─── Seed 2.0 Pro ──────────────────────────────────────────────────

\begin{table}[h]
\centering
\caption{Seed 2.0 Pro: MCQ cross-setting outcome table ($n=249$).}
\label{tab:mcq_confusion_seed}
\resizebox{\columnwidth}{!}{%
\small
\setlength{\tabcolsep}{5pt}
\begin{tabular}{@{}lrrr@{}}
\toprule
 & \textbf{Vid.\ Correct} & \textbf{Vid.\ Incorrect} & \textbf{Total} \\
\midrule
\textbf{Blind Correct}   & 65  & 40  & 105 \\
\textbf{Blind Incorrect}  & 65  & 79  & 144 \\
\midrule
\textbf{Total}            & 130 & 119 & 249 \\
\bottomrule
\end{tabular}%
}
\end{table}

\begin{table}[h]
\centering
\caption{Seed 2.0 Pro: Free-response cross-setting outcome table ($n=265$).}
\label{tab:fr_confusion_seed}
\resizebox{\columnwidth}{!}{%
\small
\setlength{\tabcolsep}{3pt}
\begin{tabular}{@{}lrrrrr@{}}
\toprule
 & \textbf{2\,pt} & \textbf{1\,pt\,(W.C.)} & \textbf{1\,pt\,(W.V.)} & \textbf{0\,pt} & \textbf{Total} \\
\midrule
\textbf{2\,pt}          & 13 & 1  & 0  & 5   & 19  \\
\textbf{1\,pt (W.C.)}   & 7  & 4  & 1  & 9   & 21  \\
\textbf{1\,pt (W.V.)}   & 10 & 1  & 4  & 8   & 23  \\
\textbf{0\,pt}          & 50 & 18 & 23 & 111 & 202 \\
\midrule
\textbf{Total}          & 80 & 24 & 28 & 133 & 265 \\
\bottomrule
\end{tabular}%
}
\end{table}

\end{document}